\documentclass[sigconf]{acmart}

\AtBeginDocument{
	}

\setcopyright{acmcopyright}
\copyrightyear{2024}
\acmYear{2024}
\acmDOI{XXXXXXX.XXXXXXX}

\acmConference[KDD '24]{the 30th ACM SIGKDD Conference on Knowledge Discovery and Data Mining}{August 25--29,
	2024}{Barcelona, Spain}
\acmISBN{978-1-4503-XXXX-X/18/06}

\usepackage{amsmath,amsfonts,bm}

\newcommand{\newterm}[1]{{\bf #1}}

\def\eqref#1{equation~\ref{#1}}

\def\1{\bm{1}}

\def\rvh{{\mathbf{h}}}

\def\vx{{\bm{x}}}
\def\vy{{\bm{y}}}

\def\mH{{\bm{H}}}

\def\mM{{\bm{M}}}

\def\mW{{\bm{W}}}
\def\mX{{\bm{X}}}

\DeclareMathAlphabet{\mathsfit}{\encodingdefault}{\sfdefault}{m}{sl}
\SetMathAlphabet{\mathsfit}{bold}{\encodingdefault}{\sfdefault}{bx}{n}

\def\gC{{\mathcal{C}}}

\def\gE{{\mathcal{E}}}

\def\gG{{\mathcal{G}}}

\def\gP{{\mathcal{P}}}

\def\gU{{\mathcal{U}}}
\def\gV{{\mathcal{V}}}

\newcommand{\softmax}{\mathrm{softmax}}

\usepackage{algorithm}
\usepackage{algorithmic}
\usepackage{multirow}
\usepackage{amsmath}
\usepackage{subcaption}

\usepackage{pifont}       
\usepackage{bbding}       
\usepackage{fontawesome}

\acmSubmissionID{1339}
\begin{document}
\title{RoutePlacer: An End-to-End Routability-Aware Placer with Graph Neural Network}

\author{Yunbo Hou}
\email{yunboh@stu.pku.edu.cn}
\affiliation{
  \institution{School of Software and Microelectronics, Peking University}
  \city{Beijing}
  \country{China}
  }

\author{Haoran Ye}
\email{haoran-ye@outlook.com}
\affiliation{
  \institution{National Key Laboratory of General Artificial Intelligence, School of Intelligence Science and Technology, Peking University}
  \city{Beijing}
  \country{China}
}

\author{Yingxue Zhang}
\email{yingxue.zhang@huawei.com}
\affiliation{
 \institution{Huawei Noah’s Ark Lab}
 \city{Beijing}
 \country{China}
 }

\author{Siyuan Xu}
\email{xusiyuan520@huawei.com}
\affiliation{
  \institution{Huawei Noah’s Ark Lab}
  \city{Beijing}
  \country{China}
  }

\author{Guojie Song}
\email{gjsong@pku.edu.cn}
\affiliation{
  \institution{National Key Laboratory of General Artificial Intelligence, School of Intelligence Science and Technology, Peking University}
  \city{Beijing}
  \country{China}
  }

\begin{abstract}

Placement is a critical and challenging step of modern chip design, with routability being an essential indicator of placement quality. Current routability-oriented placers typically apply an iterative two-stage approach, wherein the first stage generates a placement solution, and the second stage provides non-differentiable routing results to heuristically improve the solution quality. This method hinders jointly optimizing the routability aspect during placement. To address this problem, this work introduces \textbf{RoutePlacer}, an end-to-end routability-aware placement method. It trains \textbf{RouteGNN}, a customized graph neural network, to efficiently and accurately predict routability by capturing and fusing geometric and topological representations of placements. Well-trained RouteGNN then serves as a differentiable approximation of routability, enabling end-to-end gradient-based routability optimization. In addition, RouteGNN can improve two-stage placers as a plug-and-play alternative to external routers. Our experiments on DREAMPlace, an open-source AI4EDA platform, show that RoutePlacer can reduce Total Overflow by up to 16\% while maintaining routed wirelength, compared to the state-of-the-art; integrating RouteGNN within two-stage placers leads to a 44\% reduction in Total Overflow without compromising wirelength.
\end{abstract}

\maketitle

\section{Introduction}
\label{sec:intro}
The development of integrated circuits (ICs) has significantly advanced technology, progressing from individual chips to complex computing systems. Placement, among others, plays a crucial role in the intricate design flow of circuits. It constructs geometric positions of electronic components, such as memory components and logical gates, based on topological netlists. Placement can provide informative feedback for the preceding design stages and has a profound effect on downstream steps, as the positions of electronic components significantly influence the circuit performance.

Formally, the placement problem can be expressed as follows: Consider a circuit design represented by a hypergraph $H=(V,E)$, where $V$ represents the set of electronic units or cells, and $E$ represents the set of hyperedges or nets between these cells. The primary goal is to determine $\vx$ and $\vy$ to minimize wirelength while avoiding overlap between cells, where $\vx$ and $\vy$ denote cell positions.

The concept of routability is crucial when evaluating the placements of very-large-scale integrated (VLSI) circuits. Routability measures how effectively electrical signal pathways can be created on a chip. Imagine placement as deciding the locations of buildings within a city. Routing involves designing a road network for the city. The aim is to link different sections of the city (which are cells), with roads (which are wires), while preventing excessive crowding (to prevent signal interference) and fitting within the available space (complying with the chip's physical and technological limitations, such as wire width and layer spacing). Overflow, which reflects the density of wires, is the key indicator of routability. A lower overflow value indicates better routability, meaning that the chip can more efficiently form pathways that adhere to design requirements.

State-of-the-art (SOTA) analytical placers treat the placement of cells as a nonlinear optimization problem \cite{ntuplace, simpl, eplace}. Their goal is to minimize a differentiable objective function that includes the wirelength and overlap, using gradient-based optimization techniques. However, as circuits become more complex, simply focusing on this objective can lead to poor routability and routing detour failures. To address this, modern placement methods incorporate additional algorithmic modules and create an iterative two-stage process \cite{dreamplace2}: the first stage generates a placement solution, and the second stage provides non-differentiable routing results to heuristically perturb the solution. However, since these two stages are isolated and routability information is non-differentiable, these methods cannot optimize routability while generating analytical solutions, which limits the ability to jointly optimize the routability metric during placement.

To address this problem, we propose \textbf{RoutePlacer}, an end-to-end routability-aware placer that incorporates a differentiable congestion penalty into its objective function. We parameterize the congestion penalty with a graph neural network (GNN) and train it to accurately predict the congestion. The well-trained GNN can provide gradients of predicted congestion w.r.t. cell positions. This gradient information can be directly employed for gradient-based optimization, via forward and backward propagation, to minimize congestion when generating analytical placements.

During forward propagation, accurate congestion estimation is critical for reasonable penalty (gradient) assignment. To achieve this, we parameterize the congestion penalty with graph neural networks (GNNs), which have demonstrated exceptional performance in related tasks \cite{swyang-congestion}. Preserving circuit information has been highlighted as a key factor in GNN performance \cite{swyang-congestion, lhnn, crossgraph}. Therefore, we convert the circuit design, hypergraph $H=(V,E)$, into \textbf{RouteGraph}, a heterogeneous graph that preserves two sources of information: topological information from hypergraph $H=(V,E)$ and geometrical information from cell location $(\vx,\vy)$. The graph construction has a linear time complexity w.r.t. the scale of the circuit design, which ensures its efficiency. The RouteGraph contains three types of vertices: cells, nets, and grid cells. We use pins $\gP$ that connect cells and nets to represent topology (named $topo-edges$). For the geometry representation, we divide the layout into grids, each representing a grid cell. Neighboring grid cells are linked by $geom-edges$. Finally, each cell is connected to the grid cell corresponding to its location, represented by $grid-edges$.

Then, we design \textbf{RouteGNN} to give accurate routability estimation conditional on RouteGraph. To collect and enrich topological and geometrical information, RouteGNN performs message-passing on $topo-edges$, $geom-edges$, and $grid-edges$ individually and fuse the message to update representations of cells, nets, and grid cells. Multiple layers of message-passing and fusing are stacked to capture the deep relationships between topology and geometry. We sum up the routability estimation for all cells as a congestion penalty.

In backward propagation, we compute the gradient of congestion penalty w.r.t. cell locations for position updates. We introduce \textbf{Differentiable Geometrical Feature Computation} for computing the gradient of cell features w.r.t. cell locations. Employing the chain rule, we can acquire the gradient of the congestion penalty w.r.t. cell locations. Cell positions are updated using the Nesterov Accelerated Gradient (NAG) optimizer \cite{eplace}.

Interleaving forward and backward propagation yields our end-to-end routability-aware RoutePlacer. In addition, RouteGNN can improve traditional two-stage methods in a plug-and-play manner. One can replace any external router with RouteGNN to leverage its congestion estimation for routability-aware placement.

We summarize our contributions as follows.
\begin{enumerate}
\item We propose \textbf{RoutePlacer}, a routability-aware circuit placement method. It parameterizes a congestion penalty with GNN and integrates the differentiable penalty term into the optimization objective. It enables end-to-end routability optimization and improves two-stage placers in a plug-and-play manner.
\item We introduce \textbf{RouteGNN} to learn accurate routability estimations conditional on \textbf{RouteGraph}, an efficient heterogeneous graph structure with topological and geometrical features.
\item We present \textbf{Differentiable Geometrical Feature Computation} to enable gradient-based optimization. It calculates the gradient of cell features w.r.t. cell locations, preserving the complete gradient flow during backward propagation.
\item We evaluate RoutePlacer on \texttt{DAC2012} and \texttt{ISPD2011} benchmarks, based on DREAMPlace, an open-source EDA framework. RoutePlacer achieves a 16\% reduction in Total Overflow while maintaining routed wirelength compared to prior state-of-the-art (SOTA). Integrating RouteGNN within two-stage placers leads to a 44\% reduction in Total Overflow without compromising wirelength. They show the SOTA performance and extensibility of RoutePlacer.
\end{enumerate}

\section{Related Work}
\label{sec:related-work}
\subsection{Placement}
\label{sec:related-work-placement}

\begin{figure*}
	\centering
	\includegraphics{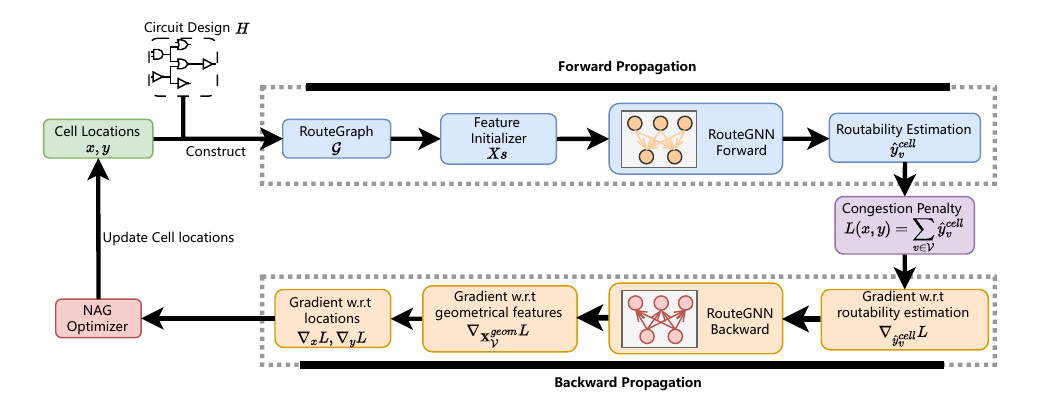}
	\caption{
 Overview of RoutePlacer. Forward Propagation: We construct the RouteGraph and initialize features, which are then inputted into RouteGNN to obtain routability estimations. $\mX s$ represents the features of cells, nets, grid cells, \emph{topo-edges}, and \emph{geom-edges}. Backward Propagation: We compute gradients of routability estimations w.r.t. cell locations via the proposed differentiable geometrical features and automatic differentiation tools. The gradient information is utilized for analytical routability optimization.}
 \Description{Forward Propagation: We construct the RouteGraph and initialize features, which are then inputted into RouteGNN to obtain routability estimations. $\mX s$ represents the features of cells, nets, grid cells, \emph{topo-edges}, and \emph{geom-edges}. Backward Propagation: We compute gradients of routability estimations w.r.t. cell locations via the proposed differentiable geometrical features and automatic differentiation tools. The gradient information is utilized for analytical routability optimization.}
	\label{fig:overview}
\end{figure*} 

Prior placement methods can mainly be divided into four categories of methods based on their optimization strategies: \newterm{Meta-heuristic} methods \cite{mdptree, hidap, ye2024llm-as-hh}, \newterm{Reinforcement Learning (RL)} methods \cite{googleplace, joint, morl, chipformer, macro-mask, kim2023devformer}, \newterm{partition-based} methods \cite{min-cut}, \newterm{Quadratic Analytical} methods \cite{ripple, utplacef, nsplace}, and \newterm{Nonlinear Analytical} methods \cite{replace, dreamplace, elfplace}. Meta-heuristic methods treat the placement as a step-wise optimization problem and solve it with a heuristic algorithm such as Simulated Annealing and Genetic Algorithm, which can theoretically reach an optimal solution. RL methods regard the placement problem as a ``board game'' and train an agent to place the cells one by one. However, these methods suffer from slow convergence, which restricts their usage only to the circuits with a small number of large-sized cells. Partition-based methods iteratively divide the chip's netlist and layout into smaller sub-netlists and sub-layouts, based on the cost function of the cut edges. Optimization methods are used to find solutions when the netlist and layout are sufficiently small.

The analytical methods are the mainstream choice for VLSI due to their efficiency and scalability. This methodological group formulates an objective function that includes wirelength and overlaps to optimize the positions of mixed-size cells. It can be further categorized into two types: quadratic and nonlinear methods. Quadratic methods \cite{ripple, utplacef, nsplace} aim to minimize wirelength and address overlaps in an alternating manner, whereas nonlinear methods \cite{replace, dreamplace, elfplace} employ a unified objective function that encompasses both wirelength and a parameterized overlap function. However, previous analytical placement approaches often overlook routability when optimizing the objective function, RoutePlacer introduces a differentiable congestion penalty into the objective function, enabling direct and synergistic routability optimization.

\subsection{Routability Optimization}

\label{sec:related-work-routability-optimization}

In modern placement, since traditional analytical placement methods struggle to guarantee satisfactory routability, two-stage routability optimization methods \cite{ntuplace4h,ntuplace4dr,simplr,crisp} have been extensively developed as coarse-grained solutions. These methods typically comprise two phases: placement and routing. The placement phase employs a traditional placement approach, while the routing stage often uses heuristics and expert-designed routers to provide feedback for the placement phase to enhance routability. Specifically, two-stage methods expand all cell sizes to make more space for coarse-grained optimization. Therefore, we propose RoutePlacer, an end-to-end placer designed for fine-grained level optimization. RoutePlacer employs gradient-based optimization for cell position updates and can integrate with two-stage methods to achieve comprehensive optimization at both coarse-grained and fine-grained levels.

\section{Proposed Approach}
\label{sec:method}

RoutePlacer is schematically illustrated in Fig. \ref{fig:overview}. We integrate a deep learning-based routability penalty into the objective function and calculate the congestion gradient to analytically optimize cell locations. Each optimization iteration consists of two steps: \textbf{Forward Propagation} and \textbf{Backward Propagation}. The forward propagation first constructs RouteGraph with its raw features. Then we input the RouteGraph and raw features into a well-trained RouteGNN to obtain routability estimation $\hat{y}^{cell}_{v}$ and sum up the estimation as routability penalty $L(\vx,\vy)=\sum_{v\in \gV}\hat{y}^{cell}_{v}$. The backward propagation first calculates the gradient of congestion penalty over raw features and the gradient of raw features over cell locations. Based on chain rules, we can compute the gradient of routability penalty over cell locations. Finally, we apply the widely used gradient-based optimization method, NAG optimizer, to update cell locations and optimize routability. Note that the optimization targets cell locations rather than GNN parameters. The RouteGNN parameters, after training, are frozen during placement optimization.
\subsection{Forward Propagation}
\label{sec:forward}
\begin{figure*}[!htbp]
	\includegraphics{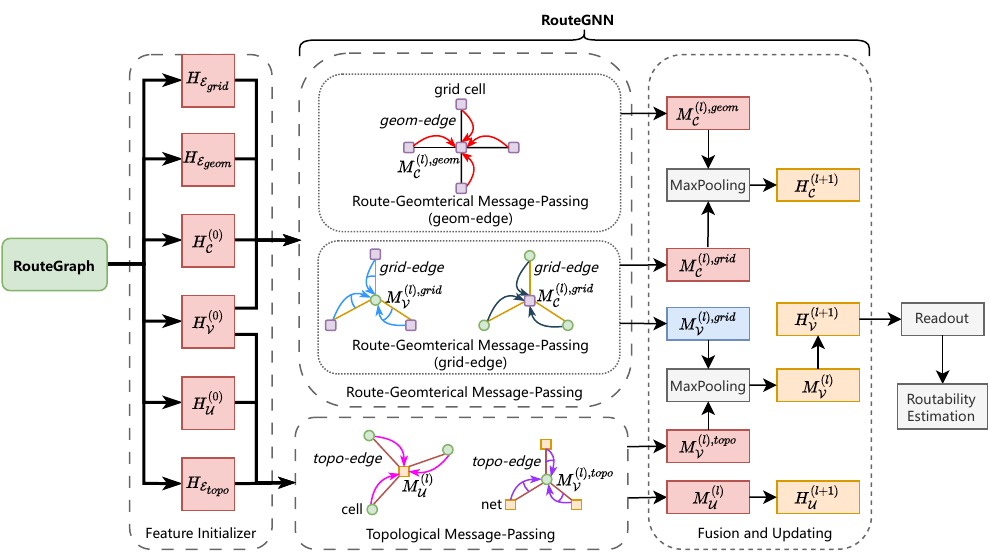}
	\caption{Framework of RouteGNN. We initialize raw features and apply Route-Geometrical and Topological message-passing to gather topological and geometrical information. Then, we fuse and update the heterogeneous hidden representations to readout routability estimation.}
 \Description{}
	\label{fig:routegnn}
\end{figure*}
\subsubsection{Circuit Design}
\label{sec:design}

The circuit design is represented as a hypergraph that stores the topological information of the circuit produced in the logic synthesis stage. The circuit design is a hypergraph $H=(\gV,\gE)$, where $\gV$ represents the set of electronic units ({cells}) and $\gE$ represents the set of hyperedges ({nets}). We transform the hypergraph into a heterogeneous graph. To construct a netlist, we consider cells $\gV$ and nets $\gE$ as two types of vertices and link each net to cells interconnected by it. We call the constructed edges \emph{topo-edges}, which stands for the topology between cells and nets.

\subsubsection{RouteGraph}
\label{sec:routegraph}

In forward propagation, we require routability estimation using graph neural networks to calculate the congestion penalty. To achieve this, it is essential to design graphs that preserve circuit information. To better retain information within a single graph during placement, we must address two key challenges: effectiveness and efficiency.

Regarding effectiveness, in chip designs, the most informative sources are circuit design and cell locations. Handling these sources independently is suboptimal. Thus, we need to integrate them into a single heterogeneous graph.

Regarding efficiency, during the placement process, cells can be distributed in a very small area. Simply connecting geometrically adjacent cells can result in a time complexity of $O(||\gV||^2)$, where $||\gV||$ represents the number of cells. This level of complexity is unacceptable when dealing with circuits that comprise millions of cells. However, restricting geometrically adjacent links to a constant number may result in the omission of numerous edges, leading to an incomplete graph structure. Therefore, a proper trade-off is required to model geometrical information with minimal loss of information.

To address these challenges, we propose RouteGraph $\gG$. For topology, we first transform the original hypergraph $H=(\gV,\gE)$ into a heterogeneous graph with two types of vertices, including cells $\gV$ and nets $\gE$. Then we link the net to the cells that the corresponding hyperedge interconnects. The edge is called \emph{topo-edges}, which stands for the topology between cells and nets.

To gather geometrical information, we begin by gridding the entire layout to create an $n \times m$ grid. Here, we set $n$ and $m$ to the numbers of routing grid cells in the horizontal and vertical directions as defined by the circuits. This grid serves as the basis for constructing a grid graph.

Within the grid graph, each node is referred to as a "grid cell" denoted as $\gC$, representing a specific grid in the layout. The edges, known as $\gE_{geom}$ or \emph{geom-edges}, signify the adjacent relationships between these grid cells. Additionally, we define \emph{grid-edges} denoted as $\gE_{grid}$ between cells and the corresponding grid cells within the grid graph. These \emph{grid-edges} represent the precise positions of cells on the layout. Furthermore, \emph{grid-edges} indirectly indicate the geometric adjacency relationships between cells.

We utilize \emph{topo-edges}, \emph{geom-edges}, and \emph{grid-edges} to effectively integrate both topological and geometrical information into RouteGraph. As the complexity of constructing each type of edge is consistently linear, the overall complexity of constructing RouteGraph remains $O(||\gP||+||\gV||+n+m)$ and is independent of cell distributions.

\subsubsection{Featurization}

\label{sec:feature}

RouteGraph comprises three types of vertices and three types of edges. In our model, \emph{geom-edges} solely represent connections between grid cells, and as such, we do not assign features to \emph{geom-edges}. We introduce $\mX_{\gV}$, $\mX_{\gU}$, $\mX_{\gC}$, $\mX_{\gE_{topo}}$, and $\mX_{\gE_{grid}}$ to denote the features of cells $\gV$, nets $\gU$, grid cells $\gC$, topological edges $\gE_{topo}$, and grid edges $\gE_{grid}$, respectively. $\mX_{\gV}$ encompasses attributes of cells $\mX_{\gV}^{attr}$, such as size and connectivity to nets, alongside the cells' geometric features $\mX_{\gV}^{geom}$, as detailed in Section \ref{sec:differentiable}. $\mX_{\gU}$ captures the span of nets as outlined in \cite{lhnn} and their connectivity to cells. $\mX_{\gC}$ incorporates features like RUDY \cite{rudy} and the central location of grid cells. $\mX_{\gE_{topo}}$ retains the intricate details of the interactions between cells and nets, employing the signal direction (input/output) as the feature for topological edges. Lastly, $\mX_{\gE_{grid}}$ records the distance between cell locations and the grid center to accurately represent their geometric adjacency.

\subsubsection{RouteGNN}

\label{sec:routegnn}

The framework of RouteGNN is shown in Fig. \ref{fig:routegnn}. RouteGNN takes a RouteGraph as input and maps raw features $\mX_{\gV}$, $\mX_{\gU}$, $\mX_{\gC}$, $\mX_{\gE_{topo}}$ and $\mX_{\gE_{grid}}$ into hidden representations $\mH_{\gV}^{(0)}$, $\mH_{\gU}^{(0)}$, $\mH_{\gC}^{(0)}$, $\mH_{\gE_{topo}}$, $\mH_{\gE_{grid}}$, via Multi-Layer Perceptrons (MLPs). Then, it generates deeper representations of cells $\gV$, nets $\gU$, and grid cells $\gC$ with $L$ layers of message-passing. Finally, the output cell representations are used for routability estimation after passing through readout layers.

In each layer, the topological information and the geometric information are collected through \textit{Topological} and \textit{Route-Geometrical} message-passing. The messages are then used to fuse and update the representations of cells $\gV$, nets $\gU$, and grid cells $\gC$. 

To collect topological information, we interact the messages of cells $\gV$ and nets $\gU$ through \emph{topo-edges} which preserve the topological connection between cells and nets. For layer $l$, we formulate the message passing as:

\begin{align}
    \mM_{\gV}^{(l),topo} & = \Phi_{msg}^{\gU \xrightarrow{\gE_{topo}} \gV}(\gU,\gE_{topo},\mH_{\gU}^{(l)},\mH_{\gE_{topo}}),  \\
	 \mM_{\gU}^{(l)} & = \Phi_{msg}^{\gV \xrightarrow{\gE_{topo}} \gU}(\gV,\gE_{topo},\mH_{\gV}^{(l)},\mH_{\gE_{topo}}), 
\end{align}
where $\mM_{\gV}^{(l),topo}$ and $\mM_{\gU}^{(l)}$ denote the messages of cells $\gV$ and nets $\gU$ computed on \emph{topo-edges} $\gE_{topo}$; $\Phi_{msg}^{\gU \xrightarrow{\gE_{topo}} \gV}$ is the message function which collects topological messages from nets $\gU$ and sends them to cells $\gV$ via \emph{topo-edges} $\gE_{topo}$. $\Phi_{msg}^{\gV \xrightarrow{\gE_{topo}} \gU}$ is similar. $\mH_{\gV}^{(l)}$ and $\mH_{\gU}^{(l)}$ denote the hidden representations of cells $\gV$ and nets $\gU$ of layer $l$. We design the message function $\Phi_{msg}^{\gV \xrightarrow{\gE_{topo}} \gU}$ as below to fuse the representations of surrounding cells ${v|(v,u)\in\gE_{topo}}$ with \emph{topo-edges} connecting them \cite{schnet}:
\begin{align}
	\Phi_{msg}^{\gV\xrightarrow{\gE_{topo}}\gU}(\{(\rvh_{v}^{\gV},\rvh_{(v,u)}^{\gE_{topo}})|(v,u)\in\gE_{topo}\})=\notag \\
	\sum_{(v,u)\in \gE_{topo}}(\mW_{\gE_{topo}\rightarrow\gU}\rvh_{(u,v)}^{\gE_{topo}}) \odot (\mW_{\gV \rightarrow \gU}\rvh_v^{\gV}),
\end{align}
where $\rvh_{v}^{\gV}$ and $\rvh_{(u,v)}^{\gE_{topo}}$ denote the hidden representation of the cell $v$ and the \emph{topo-edge} $\gE_{topo}$ connecting $v$ and $u$, respectively; $\mW_{\gE_{topo}\rightarrow\gU}$ and $\mW_{\gV \rightarrow \gU}$ are learnable weight matrices, and $\odot$ denotes the element-wise multiplication. Since the representations of \emph{topo-edges} $\mH_{\gE_{topo}}$ have been collected, we design the message function $\Phi_{msg}^{\gU \xrightarrow{\gE_{topo}} \gV}$ to speed up message-passing at the cost of minor loss of topological information. It is formally given as:

\begin{align}
	\Phi_{msg}^{\gU\xrightarrow{\gE_{topo}}\gV}(\{\rvh_{u}^{\gU}|(u,v)\in\gE_{topo}\})=
	\sum_{(u,v)\in \gE_{topo}}(\mW_{\gU \rightarrow \gV}\rvh_v^{\gU}),
\end{align}
where $\rvh_{u}^{\gU}$ is the hidden representation of net $u$ and $\mW_{\gU \rightarrow \gV}$ is a learnable weight matrix.

Geometrical information is shared via interactions between geometrically adjacent cells in RouteGraph. Initially, cell messages are fused via \emph{grid-edges}. These fused messages in grid cells are then exchanged to enable indirect interactions among adjacent cells through \emph{geom-edges} $\gE_{geom}$. Subsequently, grid cell messages are sent to cells via \emph{grid-edges} $\gE_{grid}$, enabling indirect interactions among geometrically adjacent cells.

In Route-Geometrical Message-passing, we first collect messages of cells $\gV$ to obtain a fused message of cells for grid cells $\gC$:

\begin{align}
	\mM_{\gC}^{(l), grid} = \Phi_{msg}^{\gV \xrightarrow{\gE_{grid}} \gC}(\gV,\gE_{grid},\mH_{\gV}^{(l)},\mH_{\gE_{grid}}).
\end{align}
Here, $\mM_{\gC}^{(l), grid}$ denotes the messages of grid cells $\gC$ computed on \emph{grid-edges}. $\Phi_{msg}^{\gV \xrightarrow{\gE_{grid}} \gC}$ is the message function that transfers geometrical messages from cells $\gV$ to grid cells $\gC$ via \emph{grid-edges} $\gE_{grid}$. 

Since grid cells are designed to collect geometrical messages from cells located in the grid, we design $\Phi_{msg}^{\gV \xrightarrow{\gE_{grid}} \gC}$ to speed up the message-passing as below:

\begin{align}
	&\Phi_{msg}^{\gV\xrightarrow{\gE_{grid}}\gC}(\{\rvh_{v}^{\gV}|(c,v)\in\gE_{grid}\})= 
	\sum_{(c,v)\in \gE_{grid}}(\mW_{\gV \rightarrow \gC}\rvh_v^{\gV}), 
\end{align}
where $\mW_{\gV \rightarrow \gC}$ is a learnable weight matrix.

To facilitate interaction between grid cells $\gC$ that have aggregated messages from cells $\gV$ within each grid, we utilize \emph{geom-edges} to connect these grid cells.

\begin{align}
	\mM_{\gC}^{(l),geom} = \Phi_{msg}^{\gC \xrightarrow{\gE_{geom}} \gC}(\gC,\gE_{geom},\mH_{\gC}^{(l)},\mH_{\gE_{geom}}),
\end{align}
where $\mM_{\gC}^{(l),geom}$ is the message of grid cells $\gC$ computed on \emph{geom-edges} and $\Phi_{msg}^{\gC \xrightarrow{\gE_{geom}} \gC}$ is the message function which collects geometrical messages from grid cells $\gC$;  $\mH_{\gC}^{(l)}$ denotes the hidden representations of a grid cell. Since $geom-edges$ are designed for interactions between grid cells $\gC$, we design $\Phi_{msg}^{\gC \xrightarrow{\gE_{geom}} \gC}$ to accelerate such interactions:

\begin{align}
	\Phi_{msg}^{\gC\xrightarrow{\gE_{geom}}\gC}(\{\rvh_{c}^{\gC}|(c,c^*)\in\gE_{geom}\})=
	\sum_{(c,c^*)\in \gE_{geom}}(\mW_{\gC \rightarrow \gC}\rvh_{c^*}^{\gC}), 
\end{align}
where $\rvh_{c}^{\gC}$ is the representations of grid cells $\gC$, and $\mW_{\gC \rightarrow \gC}$ is a learnable weigth matrix.

Having collected messages from geometrically adjacent grid cells, we send the grid cell messages $\gC$ to cells via \emph{grid-edges}, enabling indirect interaction among these adjacent cells.
\begin{align}
	\mM_{\gV}^{(l),grid} = \Phi_{msg}^{\gC \xrightarrow{\gE_{grid}} \gV}(\gC,\gE_{grid},\mH_{\gC}^{(l)},\mH_{\gE_{grid}}), 
\end{align}
Here, $\mM_{\gV}^{(l),grid}$ denotes the messages of cells $\gV$ computed on \emph{grid-edges}. $\Phi_{msg}^{\gC \xrightarrow{\gE_{grid}} \gV}$ is the message function that transfers messages from grid cells $\gC$ to cells $\gV$ via \emph{grid-edges} $\gE_{grid}$.

To further enhance geometrical information interaction between the cell with other surrounding cells located in the grid, we use \emph{grid-edges} $\gE_{grid}$ representations to compute edge weights when convolving cells. The message function $\Phi_{msg}^{\gC \xrightarrow{\gE_{grid}} \gV}$ is given below:

\begin{align}
	&\Phi_{msg}^{\gC\xrightarrow{\gE_{grid}}\gV}(\{(\rvh_{c}^{\gC},\rvh_{(c,v)}^{\gE_{grid}})|(c,v)\in\gE_{grid}\})=
	\notag \\ 
	&\sum_{(c,v)\in \gE_{grid}}(\alpha^T \rvh_{(c,v)}^{\gE_{grid}}) \cdot (\mW_{\gC \rightarrow \gV}\rvh_c^{\gC}), 
\end{align}
where $\rvh_{(c,v)}^{\gE_{grid}}$ is the representations vector of \emph{grid-edges} connecting $c,v$, $\alpha$ is a learnable weight vector, and $\mW_{\gC \rightarrow \gV}$ is a learnable weight matrix.

After computing topological and geometrical messages for cells $\gV$ and grid cells $\gC$ with Topological and Route-Geometrical message-passing, we fuse them and update representations for cells $\gV$ to obtain fused message $\mM_{\gV}^{(l)}$.

\begin{align}
	\mM_{\gV}^{(l)} = MaxPooling(\mM_{\gV}^{(l),grid},\mM_{\gV}^{(l),geom}).
\end{align}

The process is similar to grid cells, which aggregates cell representations from their own grids and adjacent grids through \emph{grid-edges} and \emph{geom-edges}. Consequently, we fuse the messages $\mM_{\gC}^{(l),grid}$ and $\mM_{\gC}^{(l),geom}$ to obtain the fused message $\mM_{\gC}^{(l)}$.

\begin{align}
	\mM_{\gC}^{(l)} = MaxPooling(\mM_{\gC}^{(l),grid},\mM_{\gC}^{(l),topo}).
\end{align} 

To enhance hidden representations, we fuse messages and hidden representations of layer $l$ to update hidden representations of layer $l+1$.

\begin{align}
	&\mH_{\gV}^{(l+1)} = \Phi_{update}(\mH_{\gV}^{(l)}, \mM_{\gV}^{(l)}), \notag \;\;\mH_{\gU}^{(l+1)} = \Phi_{update}(\mH_{\gU}^{(l)},\mM_{\gU}^{(l)}),  \\
	&\mH_{\gC}^{(l+1)} = \Phi_{update}(\mH_{\gC}^{(l)},\mM_{\gC}^{(l)}), 
\end{align}
where the update function $\Phi_{update}(\mH,\mM) = \mH + Tanh(\mM)$

After $L$ iterations of message-passing, we read out the cell representations $\mH_{\gV},\mH_{\gU}$ for routability estimation. For cell congestion prediction, we concatenate raw cell features with cell representations and pass them to an MLP:

\begin{align}
	\hat{y}^{cell}_v = {\rm MLP}(\mH_{\gV}^{(L)}\oplus\mX_{\gV}), 
\end{align}
where $\oplus$ is the element-wise addition. Finally, congestion penalty is formulated as $L(\vx,\vy)=\sum_{v \in \gV} \hat{y}_{v}^{cell}$.

\subsection{Backward Propagation}
\label{sec:back}

Following the computation of the congestion penalty, the next step involves backward propagating it to obtain the gradient for gradient-based optimization. Therefore, we need to calculate the gradient of the objective function w.r.t. cell locations. This section presents our approach for deriving the gradient of the congestion penalty $L$ with respect to cell locations $\nabla_x L,\nabla_y L$. As depicted in Fig. \ref{fig:overview} and guided by the chain rule, the derivatives $\nabla_x L, \nabla_y L$ are articulated as follows:

\begin{align}
	\nabla_x L= (J_x(\mX_{\gV}))^T \cdot \nabla_{\mX_{\gV}} L ,\;\;\;
	\nabla_y L= (J_y(\mX_{\gV}))^T \cdot \nabla_{\mX_{\gV}} L.
\end{align}
Here $J$ denotes a Jacobian matrix. $(J_x(\mX_{\gV}))$ and $\nabla_{\mX_{\gV}} L$ correspond to the backward process of differentiable features and \textbf{RouteGNN}. $\nabla_{\mX_{\gV} L}$ can be calculated by automatic differentiation toolkits, such as \text{PyTorch}, and $J_x(\mX_{\gV})$ is derived in Section \ref{sec:differentiable}. Finally, we apply the NAG optimizer to update cell locations using the derived gradients.

\subsubsection{Differentiable Geometrical Feature Computation}
\label{sec:differentiable}
Geometrical features play a significant role in congestion estimation \cite{rudy, swyang-congestion}. We aim to collect the gradient $J_x(\mX_{v}^{geom})$ of geometrical features over cell locations for gradient-based congestion optimization. However, geometrical features are grid features rather than cells' raw features; they are non-differentiable w.r.t. the cell positions \cite{crossgraph}. Therefore, we need to transform the grid feature into a one-dimensional vector $\mX_{\gV}^{geom}$ as cell raw features and ensure that the vector is differentiable w.r.t. cell locations. 

We propose a \textbf{Differentiable Geometrical Feature Computation} for a soft assignment of grids. The closer the routing grid is to the cell, the more wiring demand the cell has on that grid. Therefore, for each cell, we consider the closest nine grids to the cell and calculate the normalized weighted sum of RUDY, wherein the reciprocal of the distance between a cell and a grid serves the weight. The above process can be formulated as below:
\begin{align}
	& \mX_{v}^{geom}=\sum_{(a,b)\in N_v}w_{v,(a,b)}RUDY(a,b),\\
	& w_{v,(a,b)}=\softmax(\frac{1}{dis(v,(a,b))}) ,\;\;\;
	\softmax(x_i)=\frac{e^{x_i}}{\sum_ie^{x_i}},
\end{align}
where $N_v$ denotes the closest nine grids to the cell $v$, and $dis(v,(a,b))$ represents the distance between the center of the grid $(a,b)$ and the location of cells $v$. Finally, the $(J_x(\mX_{v}^{geom}))$ can be written as:
\begin{align}
	J_x(\mX_{v}^{geom})=\sum_{(a,b)\in N_v}\frac{\partial dis(v,(a,b))}{\partial x} \frac{\partial w_{v,(a,b)}}{\partial dis(v,(a,b))},
\end{align}
where $\frac{\partial dis(v,(a,b))}{\partial x}$ and $\frac{\partial w_{v,(a,b)}}{\partial dis(v,(a,b))}$ can be easily calculated in a way similar to $J_y(\mX_{v}^{geom})$.

\section{Experiments}
\label{sec:experiment}
Our experiments aim to address three research questions (RQs). (1) Effectiveness: Can RoutePlacer improve routability compared with analytical placement methods? (2) Efficiency: Can RoutePlacer efficiently handle two sources of information and give accurate routability estimation? (3) Extensibility: Can RoutePlacer improve traditional methods by replacing an external router with RouteGNN? For the first RQ, we compare RoutePlacer against DREAMPlace using routability-related metrics. For the second RQ, we compare our runtime with NetlistGNN \cite{swyang-congestion} on the collected placement and visualize the SSIM and NRMSE metrics to verify the reliability of predictions. For the third RQ, we incorporate a cell inflation method (detailed in Appendix \ref{sec:app-inflat}) into DREAMPlace and RoutePlacer, where DREAMPlace uses NCTUgr \cite{nctugr} but RoutePlacer uses RouteGNN to give feedback for cell inflation. Details of baselines and experimental settings are given in Appendix \ref{sec:app-baseline}. We conduct experiments on \texttt{ISPD2011} and \texttt{DAC2012} benchmarks, using NVIDIA RTX 3080 and Intel(R) Xeon(R) CPU E5-2699 with 31GB memory.

\subsection{Evaluating Effectiveness}
\label{sec:effectiveness}
\begin{table*}[htbp]
	\caption{Comparsion results on \texttt{ISPD2011}.}
	\label{tab:noinflat-ispd2011}
	\resizebox{1.0\linewidth}{!}{
		\begin{tabular}{lcc|cc|cc|cc|cc|cc}
			\toprule
			\multirow{2}{*}{Netlist} & \multirow{2}{*}{\#cell} & \multirow{2}{*}{\#nets} & \multicolumn{2}{c|}{TOF$\downarrow$} & \multicolumn{2}{c|}{MOF$\downarrow$} & \multicolumn{2}{c|}{H-CR$\downarrow$} & \multicolumn{2}{c|}{V-CR$\downarrow$} & \multicolumn{2}{c}{WL($\times10^6um$)$\downarrow$} \\ 
			\cmidrule{4-13} 
			& & & RoutePlacer & DREAMPlace & RoutePlacer & DREAMPlace & RoutePlacer & DREAMPlace & RoutePlacer & DREAMPlace & RoutePlacer & DREAMPlace \\ 
			\midrule
			superblue1 & 848K & 823K & \textbf{72380} & 74694 & \textbf{16} & 20 & 0.22 & 0.22 & \textbf{0.19} & 0.20 & \textbf{12.90} & 12.91 \\
			superblue2 & 1014K & 991K & \textbf{709172} & 895270 & 46 & \textbf{40} & 0.54 & \textbf{0.51} & \textbf{0.26} & 0.31 & \textbf{25.19} & 25.29 \\
			superblue4 & 600K & 568K & \textbf{114232} & 119732 & \textbf{36} & 44 & \textbf{0.45} & 0.53 & \textbf{0.21} & 0.25 & \textbf{9.18} & 9.19 \\
			superblue5 & 773K & 787K & \textbf{348976} & 363260 & \textbf{32} & 34 & 0.43 & 0.43 & 0.26 & \textbf{0.24} & \textbf{15.31} & 15.36 \\
			superblue10 & 1129K & 1086K & \textbf{142986} & 251602 & 20 & 20 & 0.29 & \textbf{0.28} & \textbf{0.15} & 0.16 & \textbf{24.10} & 24.33 \\
			superblue12 & 1293K & 1293K & \textbf{2112070} & 2201282 & \textbf{104} & 112 & \textbf{1.13} & 1.22 & 0.59 & \textbf{0.55} & \textbf{15.47} & 15.56 \\
			superblue15 & 1124K & 1080K & \textbf{115962} & 116112 & 16 & 16 & \textbf{0.25} & 0.26 & 0.13 & 0.13 & \textbf{14.51} & 14.52 \\
			superblue18 & 484K & 469K & \textbf{100336} & 104162 & 16 & 16 & \textbf{0.25} & 0.27 & 0.19 & \textbf{0.18} & \textbf{8.65} & 8.67 \\
			\multicolumn{3}{l|}{Average ratio} & 1.00 & 1.15 & 1.00 & 1.06 & 1.00 & 1.04 & 1.00 & 1.04 & 1.00 & 1.01 \\ 
			\bottomrule
		\end{tabular}
	}
\end{table*}
To evaluate the routability of placement solutions, we use NCTUgr to generate routing results and measure wirelength. We divide the entire layout into grids and assign each grid a wire limit, denoted as $RC$. This limit represents the maximum number of wires allowed in each grid cell. Overflow $OF(i,j)$ occurs when the number of wires exceeds this limit in the grid cell $(i,j)$. Wirelength refers to the total length of all wires.

Furthermore, the wire limit is categorized into two dimensions: horizontal ($RC_h$) and vertical ($RC_v$). The number of horizontal and vertical wires in each grid cell must not surpass their respective limits. $OF_h(i,j)$ and $OF_v(i,j)$ represent the excess in the number of horizontal and vertical wires, respectively.

We compare DREAMPlace with RoutePlacer based on five metrics: total overflow (TOF), wirelength (WL), maximum overflow (MOF), horizontal congestion ratio (H-CR), and vertical congestion ratio (V-CR). They are defined as follows:
\begin{align}
	& TOF=\sum_{i,j} OF(i,j) ,\;\;\;\;\;
	MOF=\max_{i,j} OF(i,j),\\
	& H-CR = \frac{\max_{i,j}OF_h(i,j)}{RC_h} ,\;\;\;\;\;
	V-CR = \frac{\max_{i,j}OF_v(i,j)}{RC_v}. 
\end{align}

The results on \texttt{ISPD2011} are shown in Table \ref{tab:noinflat-ispd2011}. We defer the results on \texttt{DAC2012} to Appendix \ref{sec:app-dac12}. Compared with DREAMPlace, RoutePlacer shows a 16\% reduction in total overflow, 2.5\% reduction in max overflow, 1\% rise in H-CR, 6\% reduction in V-CR, and maintains routed wirelength (averaged over benchmarks). The results indicate that RoutePlacer can optimize routability across various circuits.

\begin{table*}[htbp]
	\caption{Comparsion results on \texttt{ISPD2011}. RoutePlacer and DREAMPlace additionally implement cell inflation methods.}
	\label{tab:ispd11}
	\centering
	\resizebox{1.0\linewidth}{!}{
			\begin{tabular}{lcc|cc|cc|cc|cc|cc}
					\toprule
					\multirow{2}{*}{Netlist} &
					\multirow{2}{*}{\#cell} &
					\multirow{2}{*}{\#nets} &
					\multicolumn{2}{c|}{TOF$\downarrow$} &
					\multicolumn{2}{c|}{MOF$\downarrow$} &
					\multicolumn{2}{c|}{H-CR$\downarrow$} &
					\multicolumn{2}{c|}{V-CR$\downarrow$} &
					\multicolumn{2}{c}{WL($\times10^6um$)$\downarrow$} \\
					\cmidrule{4-13}
					& & & RoutePlacer & DREAMPlace & RoutePlacer & DREAMPlace & RoutePlacer & DREAMPlace & RoutePlacer & DREAMPlace & RoutePlacer & DREAMPlace \\
					\midrule
					superblue1 & 848K & 823K & \textbf{4612} & 5340 & 36 & \textbf{16} & \textbf{0.14} & 0.19 & 0.32 & \textbf{0.19} & 13.72 & \textbf{12.97} \\
					superblue2 & 1014K & 991K & \textbf{45338} & 64464 & \textbf{12} & 16 & 0.19 & \textbf{0.14} & \textbf{0.14} & 0.18 & 26.23 & \textbf{25.40} \\
					superblue4 & 600K & 568K & \textbf{6632} & 7242 & 8 & 8 & 0.18 & \textbf{0.13} & 0.13 & 0.13 & 9.79 & \textbf{9.35} \\
					superblue5 & 773K & 787K & \textbf{36726} & 108106 & \textbf{18} & 26 & \textbf{0.23} & 0.35 & 0.19 & 0.19 & 17.09 & \textbf{16.47} \\
					superblue10 & 1129K & 1086K & \textbf{44116} & 45158 & 12 & 12 & \textbf{0.20} & 0.21 & 0.13 & 0.13 & 24.91 & \textbf{24.66} \\
					superblue12 & 1293K & 1293K & \textbf{26296542} & 28714180 & \textbf{184} & 456 & 1.39 & \textbf{1.34} & \textbf{1.12} & 1.83 & \textbf{39.67} & 41.89 \\
					superblue15 & 1124K & 1080K & \textbf{15904} & 41430 & 8 & 8 & \textbf{0.18} & 0.19 & 0.09 & 0.09 & \textbf{15.07} & 15.30 \\
					superblue18 & 484K & 469K & 91388 & \textbf{22412} & 18 & \textbf{16} & 0.27 & \textbf{0.20} & 0.19 & 0.16 & \textbf{8.98} & 9.21 \\
					Average ratio& & & 1.00& 1.45 & 1.00& 1.20 & 1.00& 1.02 & 1.00& 1.04 & 1.00& 0.99 \\ 
					\bottomrule
				\end{tabular}
		}
\end{table*}

\subsection{Evaluating Efficiency}
\label{sec:efficiency}
\begin{figure}[thbp!]
	\centering
	\begin{tabular}{@{\extracolsep{\fill}}c@{}c@{\extracolsep{\fill}}}
		\includegraphics[width=0.5\linewidth]{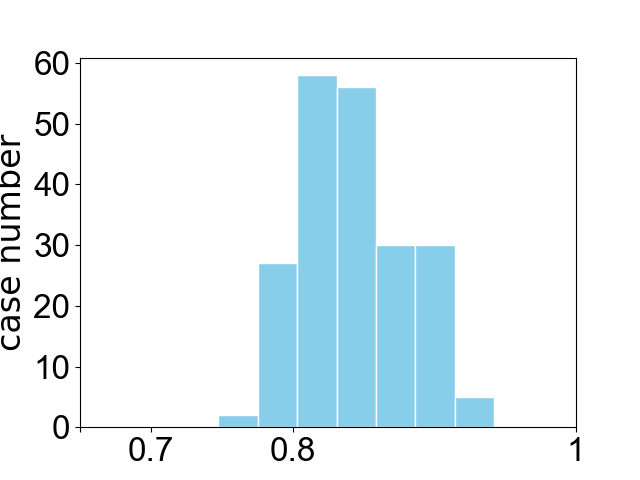} &
		\includegraphics[width=0.5\linewidth]{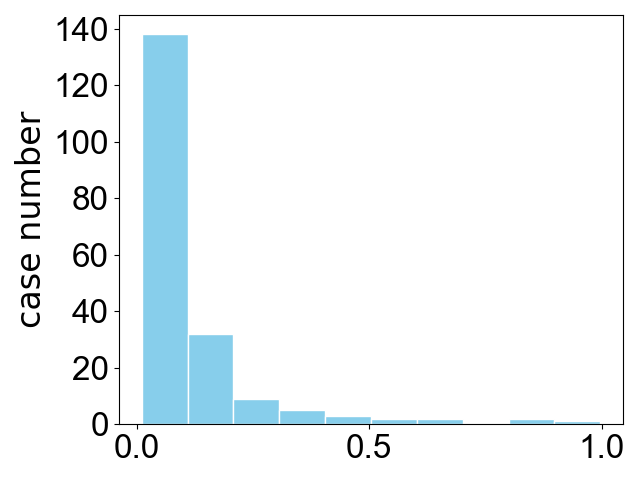}\\
		(a) SSIM statistics & (b) NRMSE statistics\\
	\end{tabular}
    \caption{Evaluations of RouteGNN. Experiments are conducted on a dataset comprising over 100 placements.}
    \Description{}
	\label{fig:estimator}
\end{figure}

\begin{figure}
	\includegraphics[width=70mm]{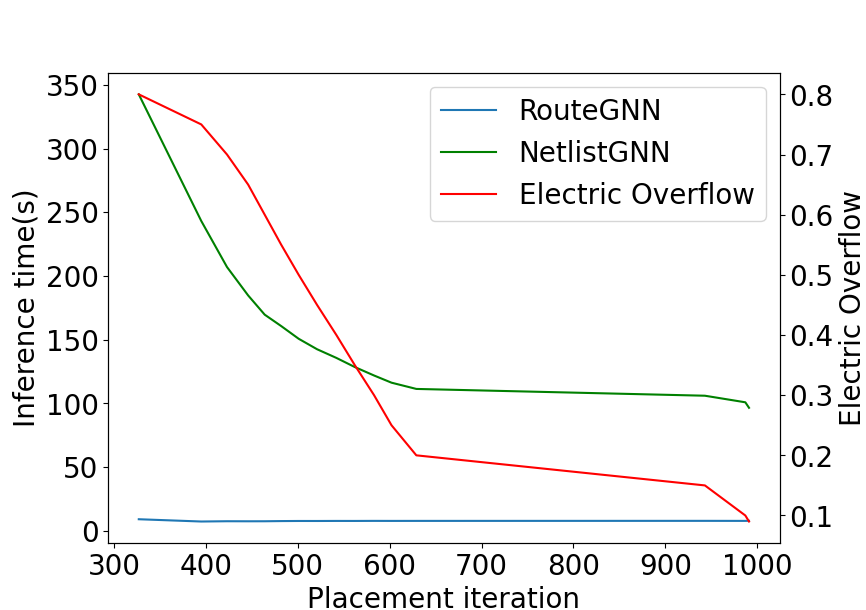}
	\caption{Comparison of inference time on \texttt{superblue7}. We additionally plot the changes of electric overflow during placement \cite{eplace}.}
 \Description{}
	\label{fig:time}
\end{figure}
To train RouteGNN, we collect placements generated by DREAMPlace and obtain labels by NCTUgr. We employ MSE loss for training, where the labels are logarithmized to prevent an excessive output range. Further details can be found in Appendix \ref{sec:app-baseline} and \ref{sec:app-train}.

To evaluate the efficiency of our model, we measure the runtime of \textbf{RouteGNN} and \textbf{NetlistGNN}. The latter connects geometric neighbors to model geometrical relationships. Fig. \ref{fig:time} visualizes the runtime comparisons between NetlistGNN and RouteGNN. We observe that as the placement process advances, cell overlap diminishes and the cells become more evenly distributed. The inference time of geometric graph construction in NetlistGNN decreases along this process. In contrast, RouteGNN consistently maintains a short runtime, unaffected by the distribution of cells.

To evaluate the accuracy of our model, we employ two key metrics: NRMSE and SSIM. NRMSE quantifies the element-wise discrepancy between our prediction and the corresponding label at the cellular level, whereas SSIM, as referenced in \cite{SSIM}, measures the structural similarity between our prediction and the ground truth at the grid level. Fig. \ref{fig:estimator} illustrates the distributions of the two statistics. In over 80\% of all cases, RouteGNN gives predictions with high accuracy, with NRMSE values below 0.1 and SSIM values above 0.8.

\subsection{Evaluating Extensibility}
\label{sec:scalablity}
\begin{figure*}
	\includegraphics{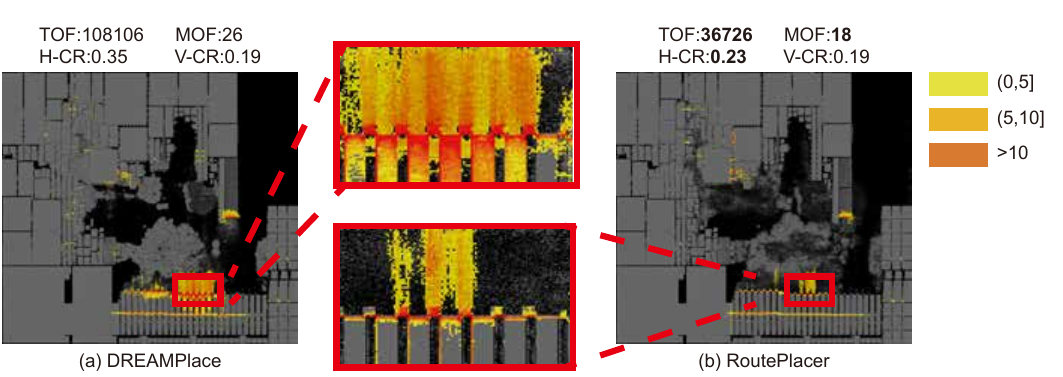}
	\caption{
 Visualization of placement solutions and their overflow on \texttt{superblue5}. The legend on the right shows the numerical range of overflow corresponding to each color. (a) A placement generated by DREAMPlace. (b) A placement generated by RoutePlacer. We implement cell inflation for both methods.
 }
 \Description{}
	\label{fig:pl}
\end{figure*}
To verify the extensibility of RoutePlacer, we incorporate a cell inflation method (detailed in Appendix \ref{sec:app-inflat}). The cell inflation method requires routing results as inputs. We use NCTUgr to generate congestion maps for DREAMPlace, while using RouteGNN to generate routability estimations for RoutePlacer. The original routability estimations of RouteGNN are on the cell level. We average the routability estimations of all cells in each grid to formulate grid maps required by the cell inflation method.
The transformation is formulated as follows:
\begin{align}
	\hat{y}^{grid}(i,j)=\frac{\sum_{v \in N_{i,j}} \hat{y}^{cell}_v}{||N_{i,j}||},
\end{align}
where $N_{i,j}$ denotes the set of cells locate in grid $i,j$, and $||N_{i,j}||$ denotes the number of cells in the set.

The results on \texttt{ISPD2011} and \texttt{DAC2012} are shown in Table \ref{tab:ispd11} and Appendix \ref{sec:app-dac12}, respectively. We report the same five metrics described above. It is demonstrated that RoutePlacer shows a 44\% reduction in total overflow, a 1\% rise in H-CR, an 8\% reduction in V-CR, and only a 1.5\% rise in routed wirelength (averaged over the benchmark). It strongly indicates that RoutePlacer can be extended to traditional two-stage pipelines. Incorporating RouteGNN into two-stage methods can yield improved routability-aware placers.

\begin{figure}
	\includegraphics[width=80mm]{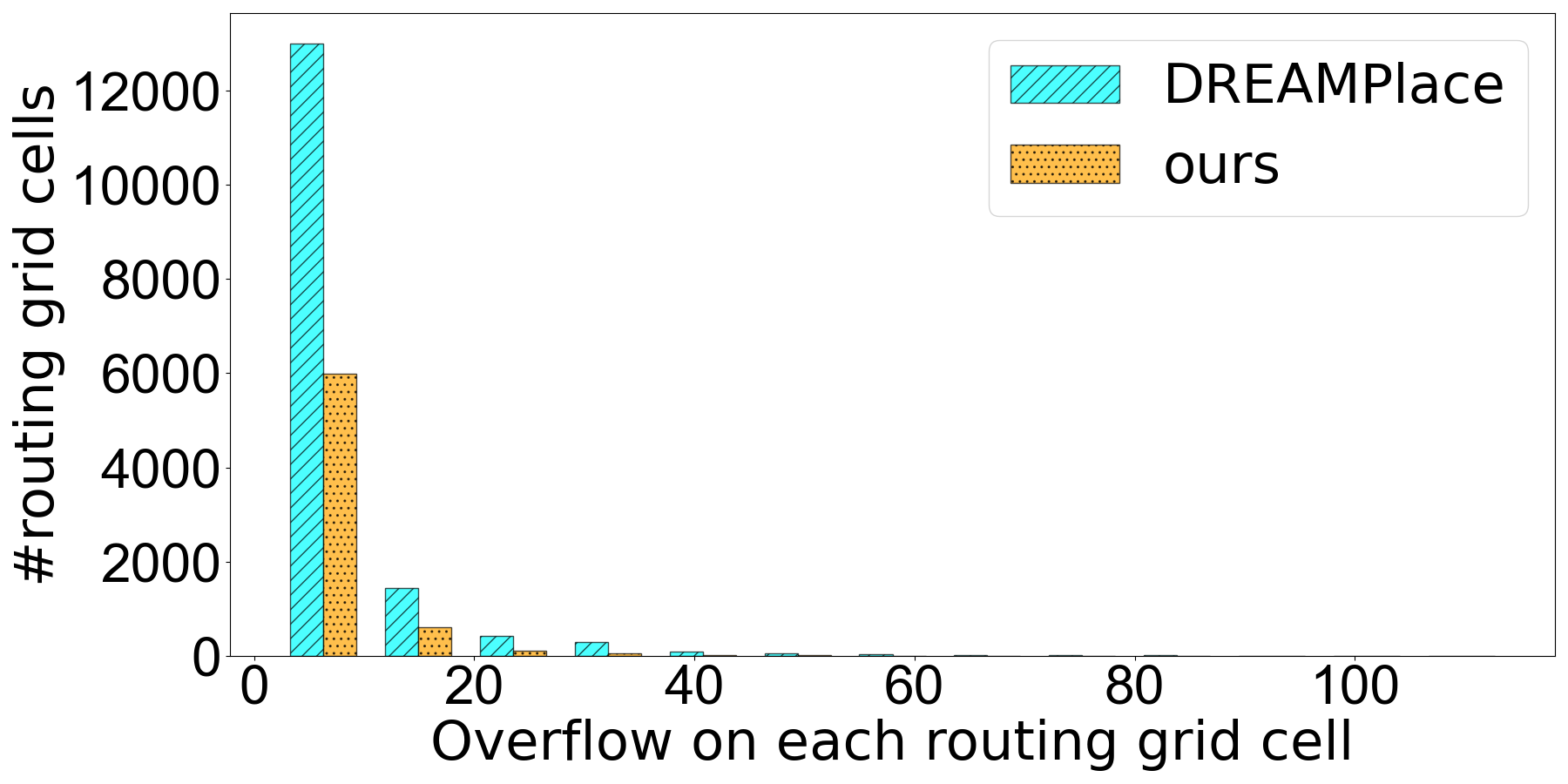}
	\caption{Overflow distributions of DREAMPlace and RoutePlacer on \texttt{superblue5}. Both of them implement cell inflation.}
 \Description{}
    \label{fig:overflow-hist}
\end{figure}

We visualize the generated placements and their overflow in Fig. \ref{fig:pl}. The cell is denoted by the grey area, and overflow visualization employs a heatmap-like method, where the intensity of the red color corresponds directly to the overflow level. The more intense the red, the higher the overflow. In Fig. \ref{fig:overflow-hist}, we present the detailed distributions of overflow. Specifically, we conduct histogram analyses of overflow $OF(i,j)$, using overflow values as the x-axis and the count of matrix elements for each value as the y-axis. These results demonstrate that RoutePlacer achieves a placement with reduced overflow compared to DREAMPlace, indicating enhanced routability.
\section{Conclusion}
\label{sec:conclusion}
This work introduces RoutePlacer, an end-to-end routability-aware placer. It enables analytical routability optimization by training RouteGNN to be a differentiable approximation of routability. Experimental results demonstrate the state-of-the-art performance of RoutePlacer in reducing overflow. RoutePlacer lays a broad foundation for future works. We plan to improve the accuracy of RouteGNN by capturing the shifts in cell positions. It is also promising to include differentiable approximations of other critical metrics, such as timing and power.

\section{ACKNOWLEDGMENTS}
\label{sec:ackowledgments}
This work was supported by the National Natural Science Foundation of China (Grant No. 62276006).

\bibliographystyle{plainnat}
\balance
\bibliography{document}

\begin{thebibliography}{31}
\providecommand{\natexlab}[1]{#1}
\providecommand{\url}[1]{\texttt{#1}}
\expandafter\ifx\csname urlstyle\endcsname\relax
  \providecommand{\doi}[1]{doi: #1}\else
  \providecommand{\doi}{doi: \begingroup \urlstyle{rm}\Url}\fi

\bibitem[A.(1977)]{min-cut}
Breuer~Melvin A.
\newblock A class of min-cut placement algorithms.
\newblock In \emph{Proceedings of the 14th Design Automation Conference}, DAC '77, page 284–290, Dakar, 1977. IEEE.
\newblock \doi{10.1145/62882.62896}.
\newblock URL \url{https://doi.org/10.1145/62882.62896}.

\bibitem[Alex et~al.(2021)Alex, Jordi, Jordi, Marc, and Ferran]{hidap}
Vidal-Obiols Alex, Cortadella Jordi, Petit Jordi, Galceran-Oms Marc, and Martorell Ferran.
\newblock Multilevel dataflow-driven macro placement guided by rtl structure and analytical methods.
\newblock \emph{IEEE Transactions on Computer-Aided Design of Integrated Circuits and Systems}, 40\penalty0 (12):\penalty0 2542--2555, Dec. 2021.
\newblock \doi{10.1109/TCAD.2020.3047724}.
\newblock URL \url{https://doi.org/10.1109/TCAD.2020.3047724}.

\bibitem[Amur et~al.(2021)Amur, Vincent, Yingxue, Dong, Wulong, and Mark]{crossgraph}
Ghose Amur, Zhang Vincent, Zhang Yingxue, Li~Dong, Liu Wulong, and Coates Mark.
\newblock Generalizable cross-graph embedding for gnn-based congestion prediction.
\newblock In \emph{2021 IEEE/ACM International Conference On Computer Aided Design}, ICCAD '21, page 1–9, Munich, Germany, 2021. IEEE.
\newblock \doi{10.1109/ICCAD51958.2021.9643446}.
\newblock URL \url{https://doi.org/10.1109/ICCAD51958.2021.9643446}.

\bibitem[Azalia and Goldie(2021)]{googleplace}
Mirhoseini Azalia and Anna Goldie.
\newblock A graph placement methodology for fast chip design.
\newblock \emph{Nature}, 594\penalty0 (7862):\penalty0 207--212, Jun. 2021.
\newblock \doi{10.1038/s41586-021-03544-w}.
\newblock URL \url{https://doi.org/10.1038/s41586-021-03544-w}.

\bibitem[Bowen et~al.(2022)Bowen, Guibao, Dong, Jianye, Wulong, Yu, Hongzhong, Yibo, Guangyong, and Ann]{lhnn}
Wang Bowen, Shen Guibao, Li~Dong, Hao Jianye, Liu Wulong, Huang Yu, Wu~Hongzhong, Lin Yibo, Chen Guangyong, and Heng~Pheng Ann.
\newblock Lhnn: lattice hypergraph neural network for vlsi congestion prediction.
\newblock In \emph{Proceedings of the 59th ACM/IEEE Design Automation Conference}, DAC '22, page 1297–1302, New York, NY, USA, 2022. Association for Computing Machinery.
\newblock \doi{10.1145/3489517.3530675}.
\newblock URL \url{https://doi.org/10.1145/3489517.3530675}.

\bibitem[Chau-Chin et~al.(2018)Chau-Chin, Hsin-Ying, Bo-Qiao, Sheng-Wei, Chin-Hao, Szu-To, Yao-Wen, Tung-Chieh, and Ismail]{ntuplace4dr}
Huang Chau-Chin, Lee Hsin-Ying, Lin Bo-Qiao, Yang Sheng-Wei, Chang Chin-Hao, Chen Szu-To, Chang Yao-Wen, Chen Tung-Chieh, and Bustany Ismail.
\newblock Ntuplace4dr: A detailed-routing-driven placer for mixed-size circuit designs with technology and region constraints.
\newblock \emph{IEEE Transactions on Computer-Aided Design of Integrated Circuits and Systems}, 37\penalty0 (3):\penalty0 669--681, Mar. 2018.
\newblock \doi{10.1109/TCAD.2017.2712665}.
\newblock URL \url{https://doi.org/10.1109/TCAD.2017.2712665}.

\bibitem[Chung-Kuan et~al.(2019)Chung-Kuan, B., Ilgweon, and Lutong]{replace}
Cheng Chung-Kuan, Kahng~Andrew B., Kang Ilgweon, and Wang Lutong.
\newblock Replace: Advancing solution quality and routability validation in global placement.
\newblock \emph{IEEE Transactions on Computer-Aided Design of Integrated Circuits and Systems}, 38\penalty0 (9):\penalty0 1717--1730, Sept. 2019.
\newblock \doi{10.1109/TCAD.2018.2859220}.
\newblock URL \url{https://doi.org/10.1109/TCAD.2018.2859220}.

\bibitem[Chung-Kuan et~al.(2022)Chung-Kuan, Chia-Tung, and Chester]{nsplace}
Cheng Chung-Kuan, Ho~Chia-Tung, and Holtz Chester.
\newblock Net separation-oriented printed circuit board placement via margin maximization.
\newblock In \emph{2022 27th Asia and South Pacific Design Automation Conference}, ASPDAC '22, page 288–293, Taipei, Taiwan, 2022. IEEE.
\newblock \doi{10.1109/ASP-DAC52403.2022.9712480}.
\newblock URL \url{https://doi.org/10.1109/ASP-DAC52403.2022.9712480}.

\bibitem[Jarrod et~al.(2009)Jarrod, Natarajan, Gi-Joon, J., and Igor]{crisp}
Roy Jarrod, Viswanathan Natarajan, Nam Gi-Joon, Alpert~Charles J., and Markov Igor.
\newblock Crisp: Congestion reduction by iterated spreading during placement.
\newblock In \emph{2009 IEEE/ACM International Conference on Computer-Aided Design}, ICCAD '09, pages 357--362, San Jose, CA, USA, 2009. IEEE.
\newblock \doi{10.1145/1687399.1687467}.
\newblock URL \url{https://doi.org/10.1145/1687399.1687467}.

\bibitem[Jingwei et~al.(2015)Jingwei, Pengwen, Chin-Chih, Lu, Jen-Hsin, Chin-Chi, and Chung-Kuan]{eplace}
Lu~Jingwei, Chen Pengwen, Chang Chin-Chih, Sha Lu, Huang~Dennis Jen-Hsin, Teng Chin-Chi, and Cheng Chung-Kuan.
\newblock eplace: Electrostatics-based placement using fast fourier transform and nesterov's method.
\newblock \emph{ACM Transactions on Design Automation of Electronic Systems}, 20\penalty0 (2):\penalty0 1--34, Mar. 2015.
\newblock \doi{10.1145/2699873}.
\newblock URL \url{https://doi.org/10.1145/2699873}.

\bibitem[Kim et~al.(2023)Kim, Kim, Berto, Kim, and Park]{kim2023devformer}
Haeyeon Kim, Minsu Kim, Federico Berto, Joungho Kim, and Jinkyoo Park.
\newblock Devformer: A symmetric transformer for context-aware device placement.
\newblock In \emph{International Conference on Machine Learning}, pages 16541--16566, Honolulu, Hawaii, USA, 2023. PMLR, JMLR.org.

\bibitem[Kristof et~al.(2018)Kristof, E., P.-J, Alexandre, and Klaus-Robert]{schnet}
Schütt Kristof, Sauceda~Huziel E., Kindermans P.-J, Tkatchenko Alexandre, and Müller Klaus-Robert.
\newblock Schnet – a deep learning architecture for molecules and materials.
\newblock \emph{The Journal of Chemical Physics}, 148\penalty0 (24):\penalty0 241722, Jun. 2018.
\newblock \doi{10.1063/1.5019779}.
\newblock URL \url{https://doi.org/10.1063/1.5019779}.

\bibitem[Lai et~al.(2023)Lai, Liu, Tang, Wang, Hao, and Luo]{chipformer}
Yao Lai, Jinxin Liu, Zhentao Tang, Bin Wang, Jianye Hao, and Ping Luo.
\newblock Chipformer: transferable chip placement via offline decision transformer.
\newblock In \emph{Proceedings of the 40th International Conference on Machine Learning}, ICML'23, page 18346–18364, Honolulu, Hawaii, USA, 2023. JMLR.org.
\newblock \doi{10.5555/3618408.3619165}.
\newblock URL \url{https://doi.org/10.5555/3618408.3619165}.

\bibitem[Meng-Kai et~al.(2014)Meng-Kai, Yi-Fang, Chau-Chin, Sheng, Tzu-Hen, Tung-Chieh, and Yao-Wen]{ntuplace4h}
Hsu Meng-Kai, Chen Yi-Fang, Huang Chau-Chin, Chou Sheng, Lin Tzu-Hen, Chen Tung-Chieh, and Chang Yao-Wen.
\newblock Ntuplace4h: A novel routability-driven placement algorithm for hierarchical mixed-size circuit designs.
\newblock \emph{IEEE Transactions on Computer-Aided Design of Integrated Circuits and Systems}, 33\penalty0 (12):\penalty0 1914--1927, Dec. 2014.
\newblock \doi{10.1109/TCAD.2014.2360453}.
\newblock URL \url{https://doi.org/10.1109/TCAD.2014.2360453}.

\bibitem[Myung-Chul et~al.(2011)Myung-Chul, Jin, Dong-Jin, and Igor]{simplr}
Kim Myung-Chul, Hu~Jin, Lee Dong-Jin, and Markov Igor.
\newblock A simplr method for routability-driven placement.
\newblock In \emph{Proceedings of the International Conference on Computer-Aided Design}, ICCAD '11, page 67–73, San Jose, California, USA, 2011. IEEE.
\newblock \doi{10.1109/ICCAD.2011.6105307}.
\newblock URL \url{https://doi.org/10.1109/ICCAD.2011.6105307}.

\bibitem[Myung-Chul et~al.(2013)Myung-Chul, Dong-Jin, and Igor]{simpl}
Kim Myung-Chul, Lee Dong-Jin, and Markov Igor.
\newblock Simpl: an algorithm for placing vlsi circuits.
\newblock \emph{Communications of the ACM}, 56\penalty0 (6):\penalty0 105–113, Jun. 2013.
\newblock \doi{10.1145/2461256.2461279}.
\newblock URL \url{https://doi.org/10.1145/2461256.2461279}.

\bibitem[Peter and Frank(2007)]{rudy}
Spindler Peter and Johannes Frank.
\newblock Fast and accurate routing demand estimation for efficient routability-driven placement.
\newblock In \emph{Proceedings of the Conference on Design, Automation and Test in Europe}, DATE '07, pages 1--6, San Jose, CA, USA, 2007. EDA Consortium.
\newblock \doi{10.1109/DATE.2007.364463}.
\newblock URL \url{https://doi.org/10.1109/DATE.2007.364463}.

\bibitem[Ruoyu and Junchi(2021)]{joint}
Cheng Ruoyu and Yan Junchi.
\newblock On joint learning for solving placement and routing in chip design.
\newblock In \emph{Advances in Neural Information Processing Systems}, NeurIPS '21, pages 16508--16519, online, 2021. Curran Associates, Inc.
\newblock \doi{10.48550/arXiv.2111.00234}.
\newblock URL \url{https://doi.org/10.48550/arXiv.2111.00234}.

\bibitem[Shi et~al.(2023)Shi, Xue, Song, and Qian]{macro-mask}
Yunqi Shi, Ke~Xue, Lei Song, and Chao Qian.
\newblock Macro placement by wire-mask-guided black-box optimization.
\newblock In \emph{Advances in Neural Information Processing Systems}, NeurIPS '23, pages 6825--6843, New Orleans, USA, 2023. Curran Associates, Inc.
\newblock \doi{10.48550/arXiv.2306.16844}.
\newblock URL \url{https://doi.org/10.48550/arXiv.2306.16844}.

\bibitem[Shuwen et~al.(2022)Shuwen, Zhihao, Dong, Yingxue, Zhanguang, Guojie, and Jianye]{swyang-congestion}
Yang Shuwen, Yang Zhihao, Li~Dong, Zhang Yingxue, Zhang Zhanguang, Song Guojie, and Hao Jianye.
\newblock Versatile multi-stage graph neural network for circuit representation.
\newblock In \emph{Advances in Neural Information Processing Systems}, NeurIPS '22, New Orleans, Louisiana, USA, 2022. Curran Associates, Inc.

\bibitem[Tung-Chieh et~al.("2005")Tung-Chieh, Tien-Chang, Zhe-Wei, and Yao-Wen"]{ntuplace}
"Chen Tung-Chieh, Hsu Tien-Chang, Jiang Zhe-Wei, and Chang Yao-Wen".
\newblock "ntuplace: a ratio partitioning based placement algorithm for large-scale mixed-size designs".
\newblock In \emph{"Proceedings of the 2005 International Symposium on Physical Design"}, "ISPD '05", page "236–238", "New York, NY, USA", "2005". "Association for Computing Machinery".
\newblock \doi{"10.1145/1055137.1055188"}.
\newblock URL \url{"https://doi.org/10.1145/1055137.1055188"}.

\bibitem[Wang et~al.(2004)Wang, Conrad, Hamid, and Eero]{SSIM}
Zhou Wang, Bovik~Alan Conrad, Sheikh Hamid, and Simoncelli Eero.
\newblock Image quality assessment: From error visibility to structural similarity.
\newblock \emph{IEEE Transactions on Image Processing}, 13\penalty0 (4):\penalty0 600--612, Apr. 2004.
\newblock \doi{10.1109/TIP.2003.819861}.
\newblock URL \url{https://doi.org/10.1109/TIP.2003.819861}.

\bibitem[Wen-Hao et~al.(2013)Wen-Hao, Wei-Chun, Yih-Lang, and Kai-Yuan]{nctugr}
Liu Wen-Hao, Kao Wei-Chun, Li~Yih-Lang, and Chao Kai-Yuan.
\newblock Nctu-gr 2.0: Multithreaded collision-aware global routing with bounded-length maze routing.
\newblock \emph{IEEE Transactions on computer-aided design of integrated circuits and systems}, 32\penalty0 (5):\penalty0 709--722, May 2013.
\newblock \doi{10.1109/TCAD.2012.2235124}.
\newblock URL \url{https://doi.org/10.1109/TCAD.2012.2235124}.

\bibitem[Wuxi et~al.(2016)Wuxi, Shounak, and Z.]{utplacef}
Li~Wuxi, Dhar Shounak, and Pan~David Z.
\newblock Utplacef: A routability-driven fpga placer with physical and congestion aware packing.
\newblock In \emph{2016 IEEE/ACM International Conference on Computer-Aided Design}, ICCAD '16, pages 869--882, Austin, TX, USA, 2016. IEEE.
\newblock \doi{10.1145/2966986.2980083}.
\newblock URL \url{https://doi.org/10.1145/2966986.2980083}.

\bibitem[Wuxi et~al.(2019)Wuxi, Yibo, and Z.]{elfplace}
Li~Wuxi, Lin Yibo, and Pan~David Z.
\newblock elfplace: Electrostatics-based placement for large-scale heterogeneous fpgas.
\newblock In \emph{2019 IEEE/ACM International Conference on Computer-Aided Design}, ICCAD '19, pages 1--8, Westminster, CO, USA, 2019. IEEE.
\newblock \doi{10.1109/ICCAD45719.2019.8942075}.
\newblock URL \url{https://doi.org/10.1109/ICCAD45719.2019.8942075}.

\bibitem[Xu et~al.(2011)Xu, Tao, Linfu, Haitong, Guxin, and Evangeline]{ripple}
He~Xu, Huang Tao, Xiao Linfu, Tian Haitong, Cui Guxin, and Young Evangeline.
\newblock Ripple: An effective routability-driven placer by iterative cell movement.
\newblock In \emph{2011 IEEE/ACM International Conference on Computer-Aided Design}, ICCAD '11, pages 74--79, San Jose, CA, USA, 2011. IEEE.
\newblock \doi{10.1109/ICCAD.2011.6105308}.
\newblock URL \url{https://doi.org/10.1109/ICCAD.2011.6105308}.

\bibitem[Yao et~al.(2022)Yao, Yao, and Ping]{morl}
Lai Yao, Mu~Yao, and Luo Ping.
\newblock Maskplace: Fast chip placement via reinforced visual representation learning.
\newblock In \emph{Advances in Neural Information Processing Systems}, NeurIPS '22, pages 24019--24030, New Orleans, Louisiana, USA, 2022. Curran Associates, Inc.
\newblock \doi{10.48550/arXiv.2211.13382}.
\newblock URL \url{https://doi.org/10.48550/arXiv.2211.13382}.

\bibitem[Ye et~al.(2024)Ye, Wang, Cao, Berto, Hua, Kim, Park, and Song]{ye2024llm-as-hh}
Haoran Ye, Jiarui Wang, Zhiguang Cao, Federico Berto, Chuanbo Hua, Haeyeon Kim, Jinkyoo Park, and Guojie Song.
\newblock Large language models as hyper-heuristics for combinatorial optimization, 2024.

\bibitem[Yen-Chun et~al.(2019)Yen-Chun, Tung-Chieh, Yao-Wen, and Sy-Yen]{mdptree}
Liu Yen-Chun, Chen Tung-Chieh, Chang Yao-Wen, and Kuo Sy-Yen.
\newblock Mdp-trees: multi-domain macro placement for ultra large-scale mixed-size designs.
\newblock In \emph{Proceedings of the 24th Asia and South Pacific Design Automation Conference}, ASPDAC '19, page 557–562, New York, NY, USA, 2019. Association for Computing Machinery.
\newblock \doi{10.1145/3287624.3287677}.
\newblock URL \url{https://doi.org/10.1145/3287624.3287677}.

\bibitem[Yibo et~al.(2019)Yibo, Shounak, Wuxi, Haoxing, Brucek, and Z.]{dreamplace}
Lin Yibo, Dhar Shounak, Li~Wuxi, Ren Haoxing, Khailany Brucek, and Pan~David Z.
\newblock Dreampiace: Deep learning toolkit-enabled gpu acceleration for modern vlsi placement.
\newblock In \emph{2019 56th ACM/IEEE Design Automation Conference}, DAC '19, pages 1--6, Las Vegas, NV, USA, 2019. IEEE.
\newblock \doi{10.1145/3316781.3317803}.
\newblock URL \url{https://doi.org/10.1145/3316781.3317803}.

\bibitem[Yibo et~al.(2020)Yibo, Z., Haoxing, and Brucek]{dreamplace2}
Lin Yibo, Pan~David Z., Ren Haoxing, and Khailany Brucek.
\newblock Dreamplace 2.0: Open-source gpu-accelerated global and detailed placement for large-scale vlsi designs.
\newblock In \emph{2020 China Semiconductor Technology International Conference}, CSTIC '20, pages 1--4, Shanghai, China, 2020. IEEE.
\newblock \doi{10.1109/CSTIC49141.2020.9282573}.
\newblock URL \url{https://doi.org/10.1109/CSTIC49141.2020.9282573}.

\end{thebibliography}

\appendix
\section{Notation}
\label{sec:app-notation}
\begin{table}[!htbp]
\begin{tabular}{ll}
\hline
Notation  & Description                                  \\ \hline
$\vx$,$\vy$       & Physical locations of cells in the layout    \\
$\gG$         & RouteGraph                                 \\
$\gV$         & Set of cells                                 \\
$\gU$         & Set of nets                                  \\
$\gC$         & Set of grid cells                            \\
\emph{topo-edge} & Set of topo-edges which denotes \\
& the topology between cells and nets            \\
\emph{geom-edge} & Set of geom-edges which signify the adjacent  \\
& relationships between these grid cells.           \\
\emph{grid-edge} & Set of grid-edges  which are denoted as edges \\
& between cells and the corresponding grid cells\\
& within the grid graph                          \\
$D(\cdot,\cdot)$         & Density Penalty                              \\
$L(\cdot,\cdot)$         & Routing Congestion Penalty                   \\
$\lambda_D$    & Weight of density penalty                    \\
$\gamma$     & Parameter of wirelength model for smoothness \\
$\eta$       & Weight of congestion penalty                 \\
 \hline
\end{tabular}
\end{table}

\section{Details of Our Method}
\label{sec:app-method}
\subsection{Loss}
Our loss function for transductive placement optimization can be expressed as 
\begin{equation}
    Loss=WL(x,y)+\lambda D(x,y)+ \eta L(x,y),
\end{equation}
where $WL(x,y)$ denotes wirelength function, $D(x,y)$ represents density function and $L(x,y)$ is routing congestion as illustrated in Section \ref{sec:method}.

\subsection{Wirelength Objective}
We follow \cite{eplace} to implement the wirelength function:
\begin{align}
    WL(x,y)=(\frac{\sum_{v\in u}{x_v}\exp\mathrm{(}\frac{x_v}{\gamma})}{\sum_{v\in u}{\exp}(\frac{x_v}{\gamma})}-\frac{\sum_{v\in u}{x_v}\exp\mathrm{(}-\frac{x_v}{\gamma})}{\sum_{v\in u}{\exp}(-\frac{x_v}{\gamma})}+ \noindent
\\
\frac{\sum_{v\in u}{x_v}\exp\mathrm{(}\frac{y_v}{\gamma})}{\sum_{v\in u}{\exp}(\frac{y_v}{\gamma})}-\frac{\sum_{v\in u}{x_v}\exp\mathrm{(}-\frac{y_v}{\gamma})}{\sum_{v\in u}{\exp}(-\frac{y_v}{\gamma})}),
\end{align}
where $x_v$ and $y_v$ denote the coordinates of cell $v$, and $\gamma$ is a hyperparameter of the wirelength model for smoothness.

\subsection{Density Objective}
\label{sec:app-density}
The unique solution of the electrostatic system (mentioned in density objective) is derived from \cite{eplace}:
\begin{align}
	&D(x,y) = \sum_{i\in v_{(x,y)}}N_i(x,y) = \sum_{i\in v_{(x,y)}} q_i\psi_i(x,y),\\
	&\left\{ \begin{array}{l}
		\nabla \cdot \nabla \psi (x,y)=-\rho (x,y)\\
		\hat{n}\cdot \psi (x,y)=\mathbf{0},(x,y)\in \partial R\\
		\iint_R{\psi}(x,y)=0\\
		\iint_R{\rho}(x,y)=0\\
	\end{array} \right. .
\end{align} 

The parameters $\lambda_D$ are updated after backward propagation and cell position updates, according to the rules given below where hyperparameters follow DREAMPlace default settings.
$$
\begin{array}{rl}
	\lambda_D&=\lambda_D*\mu, \\
	\mu &=\left\{ \begin{matrix}
		1.05*\max\mathrm{(}0.999^{epochs},0.98)&		\Delta HPWL<0\\
		1.05*1.05^{\frac{-\Delta HPWL}{350000}}&		\Delta HPWL\geqslant 0\\
	\end{matrix} \right. ,\\
	\Delta HPWL&=HPWL_{new}-HPWL_{old}.
\end{array}
$$

\subsection{Cell Inflation}
\label{sec:app-inflat}
Given a congestion map $congestion$ generated by the router, a cell inflation method expands cell size according to the rules given below when electric overflow is less than 0.2. First, for each cell, we compute the maximum congestion among grids that overlap with the cell. Then, based on the maximum congestion, we scale the cell size proportionally. The whole process is formulated as:
\begin{align}
	congestion_{exp} &= (congestion)^{exponent},\\
	increment &= \max_{(\vx,\vy) \in N_{v}} congestion_{exp}(\vx,\vy),\\
	ratio &= \sqrt{increment},\\
	size_{v}^{width} &= size_{v}^{width} * ratio, \\
	size_{v}^{height} &= size_{v}^{height} * ratio,
\end{align} 
where $exponent$ is the hyperparameters, $N_{v}$ denotes the set of grids that overlap with the cell $v$, $size_{v}^{width}$ and $size_{v}^{height}$ denote the height and width of the cell $v$.

\subsection{Training Placement Collection}
\label{sec:app-train}
To ensure diversity in the training dataset of placement, we use electric overflow \cite{eplace} as a key metric and collect placement at various electric overflow levels. This strategy allows RouteGNN to learn a richer representation of routability. Electric overflow ranging from 0 to 1 reflects the overlap level between cells. We begin collecting placements when the electric overflow first drops to 0.8; subsequently, each time the electric overflow decreases by 0.05, we collect a placement again.
\section{Baseline Settings}
\label{sec:app-baseline}

\begin{figure*}[!htbp]
	\includegraphics[width=180mm]{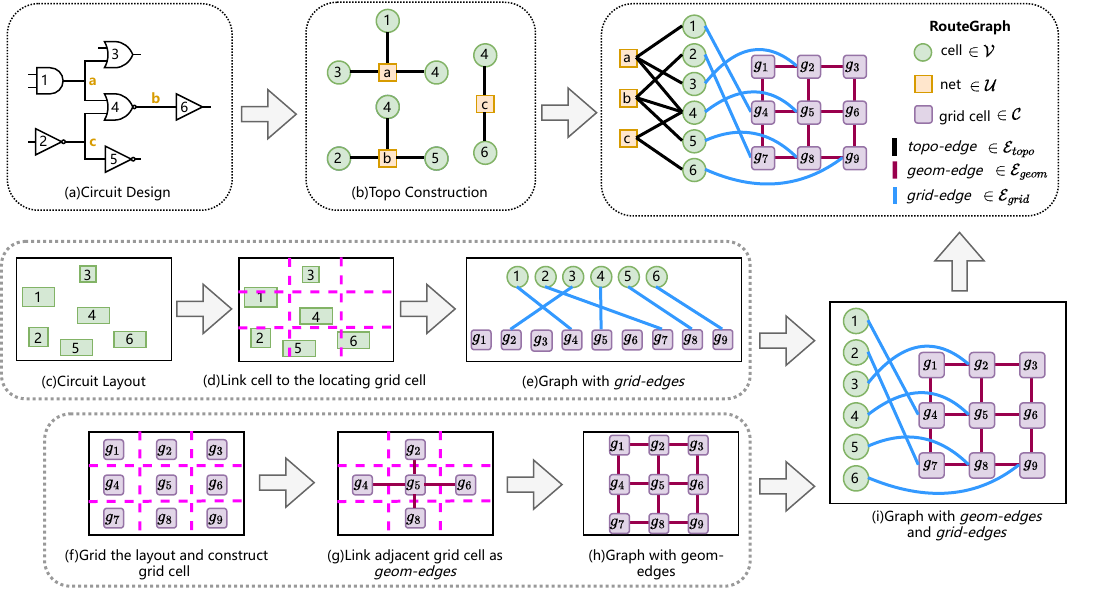}
	\caption{
 The pipeline for converting circuit design to Route Graph: (b) Build \emph{topo-edges} from cell-net connections. (f), (g), and (h): Construct grid cells and link adjacent pairs to construct \emph{geom-edges}. (c), (d), and (e): Link cells to the grid cells where they are located to construct \emph{grid-edges}.
 }
 \Description{}
	\label{fig:app-routegraph}
\end{figure*}

For RouteGNN, we set hidden layer dimensions of $cell$, $net$, $grid$ $cell$, $topo-edges$, $geom-edges$, $grid-edges$ ($F_{\gV}$, $F_{\gU}$,$F_{\gC}$,$F_{\gE_{topo}}$,$F_{\gE_{geom}}$, $,F_{\gE_{grid}})=(32,64,16,8,4,4)$ and message-passing layers $L=2$. We set the grid cell size equal to the routing grid cell size on each circuit. During training, we use Adam optimizer with learning rate $\gamma=0.0002$, learning rate decay $\Delta\gamma=0.02$, and weight decay $\eta=0.0002$; we set training epoch $e=100$ and use MSE loss for training. To avoid excessive output range, we logarithmize both label and model output. For NetlistGNN, we use its default model settings.

Both DREAMPlace and RoutePlacer use \texttt{NAG Optimizer} \cite{eplace} to ensure a fair comparison. Our training parameters are given in Table \ref{tab:param-dac12} and \ref{tab:param-ispd11}. "Num adjust" denotes the maximum number of adjustments for cell inflation. $\eta$ and $exponent$ refer to the congestion penalty and cell inflation hyperparameter, respectively, as detailed in Section \ref{sec:differentiable} and Appendix \ref{sec:app-inflat}. The adjustment strategy for $\gamma$ and $\lambda_D$, adopted by both DREAMPlace and our method, follows Lu et al. \cite{dreamplace}. For DREAMPlace and our method, the applied hyperparameters adhere to the default setting, if not mentioned in the table. Our source code is available at https://github.com/sorarain/RoutePlacer.

\begin{table}
	\centering
	\caption{Hyperparameters of our method on \texttt{DAC2012}} 
	\label{tab:param-dac12}
	\begin{tabular}{cccc}
		\toprule
		Netlist     & Num adjust & $\eta$&$exponent$      \\ \midrule
		superblue2  & 5          & 3.00e-7 & 3.5\\
		superblue3  & 6          & 1.00e-1 & 1.5\\
		superblue6  & 6          & 1.00e+1 & 2.5\\
		superblue7  & 5          & 3.00e-2 & 2.5\\
		superblue11 & 4          & 1.00e+0 & 2.5\\
		superblue12 & 4          & 1.00e+1 & 3.5\\
		superblue14 & 4          & 3.00e-2 & 2.5\\
		superblue16 & 5          & 1.00e-3 & 2.5\\
		superblue19 & 4          & 1.00e+1 & 1.5\\ \bottomrule
	\end{tabular}
\end{table}

\begin{table}
	\centering
	\caption{Hyperparameters of our method on \texttt{ISPD2011}} 
	\label{tab:param-ispd11}
	\begin{tabular}{cccc}
		\toprule
		Netlist     & Num adjust & $\eta$ &$exponent$     \\ \midrule
		superblue1  & 3          & 3.00e-2 & 2.5\\
		superblue2  & 5          & 3.00e-3 & 2.5\\
		superblue4  & 3          & 1.00e+1 & 3.5\\
		superblue5  & 5          & 1.00e+0 & 2.5\\
		superblue10 & 4          & 1.00e+1 & 2.5\\
		superblue12 & 4          & 1.00e+1 & 2.5\\
		superblue15 & 5          & 1.00e+1 & 2.5\\
		superblue18 & 5          & 3.00e-3 & 1.5\\ \bottomrule
	\end{tabular}
\end{table}

\section{Construction of RouteGraph}
\label{sec:app-routegraph}
The pipeline for constructing \textbf{RouteGraph} is depicted in Fig. \ref{fig:app-routegraph}.
\section{RUDY}
\label{sec:app-rudy}

\begin{table*}[htbp]
	\caption{Comparsion results on \texttt{DAC2012}}
	\label{tab:noinflat-dac2012}
	\resizebox{1.0\linewidth}{!}{
		\begin{tabular}{lcc|cc|cc|cc|cc|cc}
			\hline
			\multirow{2}{*}{Netlist} & \multirow{2}{*}{\#cell} & \multirow{2}{*}{\#nets} & \multicolumn{2}{c|}{TOF$\downarrow$} & \multicolumn{2}{c|}{MOF$\downarrow$} & \multicolumn{2}{c|}{H-CR$\downarrow$} & \multicolumn{2}{c|}{V-CR$\downarrow$} & \multicolumn{2}{c}{WL($\times10^6um$)$\downarrow$} \\ 
			\cmidrule{4-13} 
			& & & RoutePlacer & DREAMPlace & RoutePlacer & DREAMPlace & RoutePlacer & DREAMPlace & RoutePlacer & DREAMPlace & RoutePlacer & DREAMPlace \\ 
			\midrule
			superblue2 & 1014K & 991K & 1153398 & \textbf{1152128} & 56 & \textbf{54} & 0.56 & \textbf{0.52} & \textbf{0.27} & 0.32 & 22.83 & \textbf{22.82} \\
			superblue3 & 920K & 898K & \textbf{236218} & 240382 & 64 & \textbf{52} & 0.64 & \textbf{0.52} & \textbf{0.23} & 0.27 & 14.29 & \textbf{14.25} \\
			superblue6 & 1014K & 1007K & \textbf{132370} & 205802 & 52 & \textbf{48} & 0.51 & \textbf{0.50} & \textbf{0.22} & 0.24 & \textbf{14.33} & 14.37 \\
			superblue7 & 1365K & 1340K & \textbf{21198} & 22050 & \textbf{16} & 20 & 0.20 & \textbf{0.19} & \textbf{0.15} & 0.18 & 19.21 & \textbf{19.19} \\
			superblue11 & 955K & 936K & \textbf{46704} & 79720 & 16 & 16 & 0.22 & 0.22 & \textbf{0.12} & 0.13 & \textbf{14.36} & 14.42 \\
			superblue12 & 1293K & 1293K & 2112070 & \textbf{1478142} & 104 & \textbf{90} & 0.80 & \textbf{0.71} & 0.45 & \textbf{0.41} & 15.47 & \textbf{15.02} \\
			superblue14 & 635K & 620K & \textbf{20482} & 23490 & 24 & \textbf{20} & \textbf{0.30} & 0.25 & 0.13 & \textbf{0.12} & 9.89 & \textbf{9.88} \\
			superblue16 & 699K & 697K & \textbf{23776} & 24110 & 24 & 24 & \textbf{0.26} & 0.28 & \textbf{0.18} & 0.19 & 10.47 & \textbf{10.46} \\
			superblue19 & 523K & 512K & \textbf{61098} & 82312 & \textbf{32} & 40 & 0.34 & 0.40 & \textbf{0.18} & 0.22 & \textbf{6.72} & 6.81 \\
			\multicolumn{3}{l|}{Average ratio} & 1.00 & 1.17 & 1.00 & 0.99 & 1.00 & 0.96 & 1.00 & 1.09 & 1.00 & 1.00 \\ 
			\hline
		\end{tabular}
	}
\end{table*}

\begin{table*}[htbp]
	\caption{Comparsion results on \texttt{DAC2012}. RoutePlacer and DREAMPlace additionally implement cell inflation methods.}
	\label{tab:dac12}
	\centering
	\resizebox{1.0\linewidth}{!}{
			\begin{tabular}{lcc|cc|cc|cc|cc|cc}
					\toprule
					\multirow{2}{*}{Netlist} &
					\multirow{2}{*}{\#cell} &
					\multirow{2}{*}{\#nets} &
					\multicolumn{2}{c|}{TOF$\downarrow$} &
					\multicolumn{2}{c|}{MOF$\downarrow$} &
					\multicolumn{2}{c|}{H-CR$\downarrow$} &
					\multicolumn{2}{c|}{V-CR$\downarrow$} &
					\multicolumn{2}{c}{WL($\times10^6um$)$\downarrow$} \\
					\cmidrule{4-13}
					& & & RoutePlacer & DREAMPlace & RoutePlacer & DREAMPlace & RoutePlacer & DREAMPlace & RoutePlacer & DREAMPlace & RoutePlacer & DREAMPlace \\
					\midrule
					superblue2 & 1014K & 991K & 55690 & \textbf{43400} & \textbf{14} & 16 & 0.17 & \textbf{0.14} & \textbf{0.14} & 0.15 & 22.97 & \textbf{22.78} \\
					superblue3 & 920K & 898K & \textbf{7002} & 11064 & \textbf{8} & 14 & 0.13 & 0.13 & \textbf{0.11} & 0.14 & 15.45 & \textbf{14.65} \\
					superblue6 & 1014K & 1007K & \textbf{4880} & 5104 & 8 & 8 & \textbf{0.13} & 0.14 & \textbf{0.11} & 0.12 & 14.80 & \textbf{14.72} \\
					superblue7 & 1365K & 1340K & 17106 & \textbf{12518} & \textbf{12} & 16 & 0.17 & \textbf{0.15} & \textbf{0.13} & 0.15 & 19.65 & \textbf{19.19} \\
					superblue11 & 955K & 936K & \textbf{10140} & 19348 & \textbf{6} & 8 & 0.12 & 0.12 & \textbf{0.08} & 0.11 & 17.13 & \textbf{14.51} \\
					superblue12 & 1293K & 1293K & \textbf{3722464} & 4184918 & \textbf{114} & 116 & \textbf{0.87} & 0.88 & 0.52 & 0.52 & \textbf{20.39} & 21.73 \\
					superblue14 & 635K & 620K & 21702 & \textbf{8724} & 22 & \textbf{8} & 0.26 & \textbf{0.13} & 0.13 & \textbf{0.12} & \textbf{9.89} & \textbf{9.92} \\
					superblue16 & 699K & 697K & \textbf{14048} & 16810 & \textbf{14} & 18 & \textbf{0.18} & 0.22 & \textbf{0.13} & 0.17 & 10.59 & \textbf{10.45} \\
					superblue19 & 523K & 512K & \textbf{5278} & 22104 & 14 & 14 & \textbf{0.12} & 0.16 & 0.13 & 0.13 & 6.92 & \textbf{6.82} \\
					Average ratio& & & 1.00& 1.43 & 1.00& 1.15 & 1.00& 0.95 & 1.00& 1.13 & 1.00& 0.98 \\
					\bottomrule
				\end{tabular}
		}
\end{table*}

Rectangular uniform wire density (RUDY) \cite{rudy} provides two two-dimensional maps representing the horizontal and vertical routing demand estimations for layout bins. To compute the two estimations, we first define the bounding box of a net:
\begin{align}
	&x_e^h=\max_{p_e}x_{p_e},x_e^l=\min_{p_e}x_{p_e},y_e^h=\max_{p_e}y_{p_e},y_e^l=\max_{p_e}y_{p_e} \\
	&w_e = x_e^h - x_e^l,h_e = y_e^h - y_e^l,
\end{align}
where $x_e^l,y_e^l,x_e^h,y_e^h$ represent the left, bottom, right, and top of the bounding box of the net $e$; $p_e$ represents the pins connected to the net $e$; $x_{p_e}$ and $y_{p_e}$ represent the position of pins. Then we define the rectangle function $R(x_1^l,x_1^h,y_1^l,y_1^h;x_2^l,x_2^h,y_2^l,y_2^h)$ that calculates the overlap of two rectangle:
\begin{align}
	&w_{overlap} = min(x_1^h,x_2^h) - max(x_1^l,x_2^l),  \\
	&h_{overlap} = min(y_1^h,y_2^h) - max(y_1^l,y_2^l),  \\
	&R(x_1^l,x_1^h,y_1^l,y_1^h;x_2^l,x_2^h,y_2^l,y_2^h) = w_{overlap} * h_{overlap}, 
\end{align}

For the RUDY map in grid $\vx,\vy$, it can be calculated as follow:
\begin{align}
	{\rm RUDY}(\vx,\vy)_{h} = \sum_{e \in E} \frac{R(\vx^l,\vx^h,\vy^l,\vy^h;x_e^l,x_e^h,y_e^l,y_e^h)}{h_e},  \\
	RUDY(\vx,\vy)_{v} = \sum_{e \in E} \frac{R(\vx^l,\vx^h,\vy^l,\vy^h;x_e^l,x_e^h,y_e^l,y_e^h)}{w_e}, 
\end{align}
where $E$ represents the whole net set, and $\vx^l,\vx^h,\vy^l,\vy^h$ represent the bottom, top, left, and right edge of the grid $\vx,\vy$. $RUDY(\vx,\vy)_h$ and $RUDY(\vx,\vy)_{v}$ represent the horizontal and vertical routing demand estimation, respectively.
\section{Extended Results}
\label{sec:app-exp}

\subsection{Extended Results on \texttt{DAC2012}}
\label{sec:app-dac12}
We evaluate the effectiveness and extensibility of RoutePlacer on \texttt{DAC2012}, with results presented in Table \ref{tab:noinflat-dac2012} and Table \ref{tab:dac12}.

\begin{table*}[ht]
\caption{Extended results on \texttt{ISPD2011}.}
\label{tab:rep-ispd11-noinflat}
\resizebox{1.0\linewidth}{!}{
\begin{tabular}{lcc|cc|cc|cc|cc|ll}
\hline
\multicolumn{1}{c}{\multirow{2}{*}{Netlist}} &
  \multirow{2}{*}{\#cell$\downarrow$} &
  \multirow{2}{*}{\#nets$\downarrow$} &
  \multicolumn{2}{c|}{TOF$\downarrow$} &
  \multicolumn{2}{c|}{MOF$\downarrow$} &
  \multicolumn{2}{c|}{H-CR$\downarrow$} &
  \multicolumn{2}{c|}{V-CR$\downarrow$} &
  \multicolumn{2}{c}{WL($\times10^6um$)$\downarrow$} \\ \cline{4-13} 
\multicolumn{1}{c}{} &
   &
   &
  RoutePlacer &
  DREAMPlace &
  RoutePlacer &
  DREAMPlace &
  RoutePlacer &
  DREAMPlace &
  RoutePlacer &
  DREAMPlace &
  RoutePlacer &
  DREAMPlace \\ \hline
superblue1 &
  848K &
  823K &
  74602±1869.38 &
  75223±757.19 &
  18±0.0 &
  21±2.71 &
  0.23±0.02 &
  0.26±0.05 &
  0.19±0.0 &
  0.2±0.01 &
  12.89±0.01 &
  12.89±0.02 \\
superblue2 &
  1014K &
  991K &
  772616±13141.35 &
  915654±9258.85 &
  42±1.79 &
  40±1.6 &
  0.51±0.03 &
  0.51±0.01 &
  0.26±0.02 &
  0.29±0.03 &
  24.33±0.04 &
  25.33±0.01 \\
superblue4 &
  600K &
  568K &
  137133±32968.61 &
  119572±2728.94 &
  31±17.23 &
  34±2.53 &
  0.50±0.07 &
  0.43±0.02 &
  0.2±0.04 &
  0.23±0.02 &
  9.19±0.02 &
  9.19±0.01 \\
superblue5 &
  773K &
  787K &
  315207±2376.55 &
  356746±6727.05 &
  30±0.00 &
  34±1.6 &
  0.45±0.02 &
  0.44±0.01 &
  0.24±0.01 &
  0.24±0.02 &
  15.53±0.01 &
  15.31±0.01 \\
superblue10 &
  1129K &
  1086K &
  108494±9158.13 &
  244544±5717.01 &
  20±0.8 &
  22±3.1 &
  0.29±0.01 &
  0.31±0.03 &
  0.16±0.02 &
  0.17±0.01 &
  24.67±0.01 &
  24.29±0.02 \\
superblue12 &
  1293K &
  1293K &
  2113538±5569.93 &
  2191998±23031.14 &
  104±1.5 &
  112±7.33 &
  1.10±0.02 &
  1.19±0.08 &
  0.60±0.02 &
  0.58±0.02 &
  15.32±0.02 &
  15.57±0.01 \\
superblue15 &
  1124K &
  1080K &
  118182±3404.5 &
  112199±3565.51 &
  16±6.4 &
  16±0.0 &
  0.25±0.07 &
  0.25±0.02 &
  0.13±0.02 &
  0.13±0.0 &
  9.17±0.01 &
  14.51±0.00 \\
superblue18 &
  484K &
  469K &
  103876±8033.05 &
  106429±4790.74 &
  21±1.6 &
  22±1.96 &
  0.32±0.01 &
  0.31±0.02 &
  0.23±0.02 &
  0.18±0.01 &
  15.29±0.01 &
  8.66±0.01 \\
Average ratio &
  \multicolumn{1}{l}{} &
  \multicolumn{1}{l}{} &
  1.00 &
  1.18 &
  1.00 &
  1.07 &
  1.00 &
  1.01 &
  1.00 &
  1.02 &
  1.00 &
  1.02 \\ \hline
\end{tabular}
}
\end{table*}

\begin{table*}[ht]
\caption{Extended results on \texttt{ISPD2011}.  RoutePlacer and DREAMPlace additionally implement cell inflation methods.}
\label{tab:rep-ispd2011}
\resizebox{1.0\linewidth}{!}{
\begin{tabular}{lcc|cc|cc|cc|cc|cc}
\hline
\multicolumn{1}{c}{\multirow{2}{*}{Netlist}} &
  \multirow{2}{*}{\#cell} &
  \multirow{2}{*}{\#nets} &
  \multicolumn{2}{c|}{TOF$\downarrow$} &
  \multicolumn{2}{c|}{MOF$\downarrow$} &
  \multicolumn{2}{c|}{H-CR$\downarrow$} &
  \multicolumn{2}{c|}{V-CR$\downarrow$} &
  \multicolumn{2}{c}{WL($\times10^6um$)$\downarrow$} \\ \cline{4-13} 
\multicolumn{1}{c}{} &
   &
   &
  RoutePlacer &
  DREAMPlace &
  RoutePlacer &
  DREAMPlace &
  RoutePlacer &
  DREAMPlace &
  RoutePlacer &
  DREAMPlace &
  RoutePlacer &
  DREAMPlace \\ \hline
superblue1 &
  848K &
  823K &
  4417±375.29 &
  6050±344.88 &
  28±1.60 &
  16±2.19 &
  0.21±0.02 &
  0.22±0.02 &
  0.26±0.01 &
  0.18±0.02 &
  13.35±0.01 &
  12.97±0.00 \\
superblue2 &
  1014K &
  991K &
  31255±1977.15 &
  60975±1744.39 &
  16±1.26 &
  14±1.60 &
  0.15±0.02 &
  0.14±0.00 &
  0.17±0.02 &
  0.16±0.01 &
  26.33±0.03 &
  25.41±0.01 \\
superblue4 &
  600K &
  568K &
  6136±325.09 &
  7220±547.88 &
  9±0.94 &
  8±0.80 &
  0.17±0.02 &
  0.14±0.00 &
  0.13±0.00 &
  0.13±0.00 &
  9.64±0.17 &
  9.35±0.01 \\
superblue5 &
  773K &
  787K &
  29184±1232.17 &
  117028±411517.14 &
  12±1.60 &
  29±7.11 &
  0.21±0.01 &
  0.38±0.06 &
  0.14±0.00 &
  0.2±0.01 &
  16.00±0.02 &
  19.03±2.00 \\
superblue10 &
  1129K &
  1086K &
  46018±2381.43 &
  47606±3908.47 &
  11±0.94 &
  12±1.6 &
  0.19±0.00 &
  0.21±0.0 &
  0.13±0.00 &
  0.14±0.0 &
  23.16±0.01 &
  24.63±0.04 \\
superblue12 &
  1293K &
  1293K &
  14595543±10669254.2 &
  15112277±11283429.81 &
  159.2±52.97 &
  201±91.63 &
  1.26±0.07 &
  1.24±0.08 &
  0.93±0.38 &
  1.05±0.4 &
  29.44±9.45 &
  29.91±10.01 \\
superblue15 &
  1124K &
  1080K &
  15394±1559.95 &
  45317±59502.37 &
  8±0.80 &
  21±7.96 &
  0.18±0.01 &
  0.3±0.08 &
  0.13±0.00 &
  0.12±0.02 &
  15.00±0.00 &
  15.62±0.34 \\
superblue18 &
  484K &
  469K &
  33221±3264.33 &
  19982±6057.54 &
  18±1.50 &
  16±0.98 &
  0.22±0.04 &
  0.2±0.02 &
  0.19±0.01 &
  0.18±0.01 &
  8.81±0.01 &
  9.11±0.11 \\
Average ratio &
  \multicolumn{1}{l}{} &
  \multicolumn{1}{l}{} &
  1.00 &
  1.77 &
  1.00 &
  1.33 &
  1.00 &
  1.16 &
  1.00 &
  1.02 &
  1.00 &
  1.03 \\ \hline
\end{tabular}
}
\end{table*}

\begin{table*}[ht]
\caption{Extended results on \texttt{DAC2012}. }
\label{tab:rep-dac12-noinflat}
\resizebox{1.0\linewidth}{!}{
\begin{tabular}{lcc|cc|cc|cc|cc|cc}
\hline
\multicolumn{1}{c}{\multirow{2}{*}{Netlist}} &
  \multirow{2}{*}{\#cell} &
  \multirow{2}{*}{\#nets} &
  \multicolumn{2}{c|}{TOF$\downarrow$} &
  \multicolumn{2}{c|}{MOF$\downarrow$} &
  \multicolumn{2}{c|}{H-CR$\downarrow$} &
  \multicolumn{2}{c|}{V-CR$\downarrow$} &
  \multicolumn{2}{c}{WL($\times10^6um$)$\downarrow$} \\ \cline{4-13} 
\multicolumn{1}{c}{} &
   &
   &
  RoutePlacer &
  DREAMPlace &
  RoutePlacer &
  DREAMPlace &
  RoutePlacer &
  DREAMPlace &
  RoutePlacer &
  DREAMPlace &
  RoutePlacer &
  DREAMPlace \\ \hline
superblue2 &
  1014K &
  991K &
  1150063±6745.57 &
  1152809±16240.65 &
  57±1.6 &
  55±1.96 &
  0.57±0.01 &
  0.55±0.02 &
  0.26±0.02 &
  0.26±0.02 &
  22.84±0.04 &
  22.75±0.06 \\
superblue3 &
  920K &
  898K &
  236023±4326.52 &
  240571±2374.81 &
  56±2.83 &
  53±4.31 &
  0.55±0.03 &
  0.54±0.03 &
  0.25±0.02 &
  0.26±0.02 &
  14.27±0.01 &
  14.26±0.01 \\
superblue6 &
  1014K &
  1007K &
  138225±5877.82 &
  207011±2579.4 &
  50±2.53 &
  48±2.53 &
  0.5±0.02 &
  0.49±0.02 &
  0.24±0.02 &
  0.25±0.02 &
  14.38±0.01 &
  14.39±0.01 \\
superblue7 &
  1365K &
  1340K &
  23836±1481.67 &
  24863±1012.67 &
  17±3.67 &
  17±0.98 &
  0.2±0.02 &
  0.22±0.02 &
  0.16±0.03 &
  0.17±0.0 &
  19.21±0.01 &
  19.22±0.01 \\
superblue11 &
  955K &
  936K &
  48600±3456.19 &
  79745±5595.66 &
  17±1.6 &
  17±1.6 &
  0.22±0.01 &
  0.22±0.01 &
  0.13±0.0 &
  0.13±0.0 &
  14.38±0.00 &
  14.40±0.01 \\
superblue12 &
  1293K &
  1293K &
  1422014±29442.55 &
  1455114±21688.64 &
  88±5.12 &
  99±3.71 &
  0.69±0.04 &
  0.75±0.02 &
  0.4±0.02 &
  0.41±0.02 &
  14.93±0.01 &
  15.00±0.01 \\
superblue14 &
  635K &
  620K &
  22385±955.02 &
  24399±983.83 &
  15±2.65 &
  16±2.04 &
  0.22±0.03 &
  0.23±0.02 &
  0.12±0.0 &
  0.12±0.0 &
  9.87±0.01 &
  9.89±0.00 \\
superblue16 &
  699K &
  697K &
  23858±577.41 &
  23898±514.66 &
  22±1.5 &
  22±2.33 &
  0.24±0.01 &
  0.25±0.02 &
  0.18±0.01 &
  0.18±0.02 &
  10.45±0.01 &
  10.46±0.01 \\
superblue19 &
  523K &
  512K &
  74537±1309.32 &
  85839±1146.27 &
  36±3.79 &
  39±8.04 &
  0.35±0.03 &
  0.38±0.07 &
  0.22±0.0 &
  0.22±0.0 &
  6.80±0.00 &
  6.81±0.00 \\
Average ratio &
  \multicolumn{1}{l}{} &
  \multicolumn{1}{l}{} &
  1.00 &
  1.16 &
  1.00 &
  1.02 &
  1.00 &
  1.03 &
  1.00 &
  1.02 &
  1.00 &
  1.00 \\ \hline
\end{tabular}
}
\end{table*}

\begin{table*}[ht]
\caption{Extended results on \texttt{DAC2012}. RoutePlacer and DREAMPlace additionally implement cell inflation methods.}
\label{tab:rep-dac12}
\resizebox{1.0\linewidth}{!}{
\begin{tabular}{lcc|cc|cc|cc|cc|cc}
\hline
\multicolumn{1}{c}{\multirow{2}{*}{Netlist}} &
  \multirow{2}{*}{\#cell} &
  \multirow{2}{*}{\#nets} &
  \multicolumn{2}{c|}{TOF$\downarrow$} &
  \multicolumn{2}{c|}{MOF$\downarrow$} &
  \multicolumn{2}{c|}{H-CR$\downarrow$} &
  \multicolumn{2}{c|}{V-CR$\downarrow$} &
  \multicolumn{2}{c}{WL($\times10^6um$)$\downarrow$} \\ \cline{4-13} 
\multicolumn{1}{c}{} &
   &
   &
  RoutePlacer &
  DREAMPlace &
  RoutePlacer &
  DREAMPlace &
  RoutePlacer &
  DREAMPlace &
  RoutePlacer &
  DREAMPlace &
  RoutePlacer &
  DREAMPlace \\ \hline
superblue2 &
  1014K &
  991K &
  49762±2444.92 &
  43968±548.44 &
  15±1.60 &
  16±1.58 &
  0.18±0.01 &
  0.14±0.01 &
  0.14±0.01 &
  0.15±0.02 &
  2.30±0.02 &
  22.79±0.02 \\
superblue3 &
  920K &
  898K &
  10988±3483.21 &
  11284±735.63 &
  10±2.30 &
  12±1.26 &
  0.13±0.00 &
  0.13±0.01 &
  0.13±0.00 &
  0.13±0.01 &
  15.58±0.18 &
  14.71±0.01 \\
superblue6 &
  1014K &
  1007K &
  4677±762.03 &
  4831±375.02 &
  8±1.60 &
  8±0.00 &
  0.15±0.02 &
  0.16±0.01 &
  0.1±0.01 &
  0.11±0.01 &
  15.92±0.02 &
  14.69±0.03 \\
superblue7 &
  1365K &
  1340K &
  14440±965.37 &
  11648±339.91 &
  16±3.25 &
  15±0.8 &
  0.18±0.02 &
  0.15±0.02 &
  0.16±0.02 &
  0.16±0.01 &
  20.00±0.01 &
  19.20±0.01 \\
superblue11 &
  955K &
  936K &
  18695±4151.57 &
  19377±2914.19 &
  8±0.80 &
  9±0.98 &
  0.16±0.00 &
  0.12±0.00 &
  0.11±0.0 &
  0.12±0.00 &
  17.20±0.03 &
  14.50±0.03 \\
superblue12 &
  1293K &
  1293K &
  3779873±116464.08 &
  4112154±113047.13 &
  106±2.82 &
  118±1.88 &
  0.83±0.04 &
  0.89±0.04 &
  0.52±0.01 &
  0.52±0.01 &
  20.54±0.11 &
  21.31±0.47 \\
superblue14 &
  635K &
  620K &
  21423±3146.30 &
  6793±967.36 &
  20±4.12 &
  11±2.99 &
  0.26±0.03 &
  0.13±0.01 &
  0.13±0.0 &
  0.12±0.01 &
  10.19±0.00 &
  9.92±0.01 \\
superblue16 &
  699K &
  697K &
  12313±545.20 &
  16800±481.60 &
  16±0.98 &
  16±0.00 &
  0.21±0.01 &
  0.18±0.02 &
  0.14±0.01 &
  0.16±0.00 &
  10.77±0.00 &
  10.45±0.01 \\
superblue19 &
  523K &
  512K &
  5132±413.30 &
  20916±997.21 &
  10±0.98 &
  15±0.98 &
  0.12±0.00 &
  0.16±0.00 &
  0.12±0.0 &
  0.14±0.01 &
  6.93±0.01 &
  6.81±0.01 \\
Average ratio &
  \multicolumn{1}{l}{} &
  \multicolumn{1}{l|}{} &
  1.00 &
  1.29 &
  1.00 &
  1.05 &
  1.00 &
  0.91 &
  1.00 &
  1.05 &
  1.00 &
  0.96 \\
  \hline
\end{tabular}
}
\end{table*}

\subsection{Sensitivity to Message Layer Hyperparameter}
\label{sec:app-sens-layer}
We evaluate the hyperparameter sensitivity of RouteGNN to the number of message layers. The results are collected in Table \ref{tab:sens-layer}. We find that $L=2$ gives the best trade-off between prediction accuracy and model complexity, and RouteGNN is generally not sensitive to $L$.

\begin{table}[H]
	\caption{Sensitivity of RouteGNN to the number of message layers. We test on \texttt{superblue1/2/3/4/5/6/7/10/11/12/14/15}. The results are averaged across netlists on the cell level.}
	\label{tab:sens-layer}
	\begin{tabular}{cccccc}
		\toprule
		\# of Layers & pearson       & spearman      & kendall & time  & time ratio \\ \midrule
		1            & 0.62          & 0.64          & 0.53    & 11.80 & 0.92       \\
		2            & \textbf{0.64} & 0.65          & 0.53    & 12.87 & 1.00       \\
		3            & \textbf{0.64} & \textbf{0.66} & 0.53    & 13.91 & 1.08       \\ \bottomrule
	\end{tabular}
\end{table}

\subsection{Sensitivity to Placement Hyperparmeters }
\label{sec:app-hyper}

\begin{table*}[ht]
    \caption{Ablation study of RouteGNN on \texttt{DAC2012}. "Trained": RoutePlacer with well-trained RouteGNN. "Random": RoutePlacer with randomly parameterized RouteGNN. "-": the placement results that fail to route}
    \label{tab:dac12-train-random}
    \resizebox{1.0\linewidth}{!}{
    \begin{tabular}{ccc|cc|cc|cc|cc|cc}
        \hline
        \multirow{2}{*}{Netlist} &
        \multirow{2}{*}{\#cell} &
        \multirow{2}{*}{\#nets} &
        \multicolumn{2}{c|}{TOF$\downarrow$} &
        \multicolumn{2}{c|}{MOF$\downarrow$} &
        \multicolumn{2}{c|}{H-CR$\downarrow$} &
        \multicolumn{2}{c|}{V-CR$\downarrow$} &
        \multicolumn{2}{c}{WL ($\times 10^6 um$)↓} \\
        \cmidrule{4-13}
        &
        &
        &
        \multicolumn{1}{l}{Trained} &
        \multicolumn{1}{l|}{Random} &
        \multicolumn{1}{l}{Trained} &
        \multicolumn{1}{l|}{Random} &
        \multicolumn{1}{l}{Trained} &
        \multicolumn{1}{l|}{Random} &
        \multicolumn{1}{l}{Trained} &
        \multicolumn{1}{l|}{Random} &
        \multicolumn{1}{l}{Trained} &
        \multicolumn{1}{l}{Random} \\
        \midrule
        superblue2  & 1014K & 991K  & \textbf{55690}   & -       & \textbf{14}   & -    & \textbf{0.17} & -    & \textbf{0.14} & -    & \textbf{22.97} & -              \\
        superblue3  & 920K  & 898K  & \textbf{7002}    & 1703046 & \textbf{8}    & 56   & \textbf{0.13} & 0.57 & \textbf{0.11} & 0.27 & \textbf{15.45} & 20.41          \\
        superblue6  & 1014K & 1007K & \textbf{4880}    & 240052  & \textbf{8}    & 52   & \textbf{0.13} & 0.51 & \textbf{0.11} & 0.27 & \textbf{14.80} & 15.56          \\
        superblue7  & 1365K & 1340K & \textbf{17106}   & 324060  & \textbf{12}   & 24   & \textbf{0.17} & 0.30 & \textbf{0.13} & 0.19 & \textbf{19.65} & 25.55          \\
        superblue11 & 955K  & 936K  & \textbf{10140}   & 151552  & \textbf{6}    & 24   & \textbf{0.12} & 0.28 & \textbf{0.08} & 0.13 & 17.13          & \textbf{16.80} \\
        superblue12 & 1293K & 1293K & \textbf{3722464} & 4271624 & \textbf{114}  & 124  & \textbf{0.87} & 0.94 & 0.52          & 0.52 & \textbf{20.39} & 21.83          \\
        superblue14 & 635K  & 620K  & \textbf{21702}   & 41402   & \textbf{22}   & 26   & \textbf{0.26} & 0.31 & \textbf{0.13} & 0.17 & \textbf{9.89}  & 10.23          \\
        superblue16 & 699K  & 697K  & \textbf{14048}   & 50310   & \textbf{14}   & 16   & \textbf{0.18} & 0.21 & \textbf{0.13} & 0.14 & \textbf{10.59} & 11.62          \\
        superblue19 & 523K  & 512K  & \textbf{7284}    & 350142  & \textbf{12}   & 72   & \textbf{0.15} & 0.41 & \textbf{0.13} & 0.50 & \textbf{6.99}  & 8.31           \\
        Average ratio&       &       & 1.00             & 42.45   & {1.00} & 3.32 & {1.00} & 2.18 & 1.00 & 1.80 & 1.00  & 1.12      \\
        \hline    
    \end{tabular}
}
\end{table*}

We evaluate the sensitivity to placement hyperparameters $\eta$ and "Num adjust", with results gathered in Table \ref{tab:sens-eta}, \ref{tab:sens-adjust} and \ref{tab:sens-exp}. We find that, on \texttt{superblue19}, setting $\eta=10$, $exponent=1.5$, and "Num adjust" equal to 5 is the best hyperparameter choice. Note that "Num adjust" is introduced in Appendix \ref{sec:app-baseline}.

\begin{table}[H]
	\caption{Sensivity of $\eta$ (tested on \texttt{superblue19} with no use of cell inflation). TOF represents the sum of $OF(i,j,k)$}
	\label{tab:sens-eta}
	\begin{tabular}{ccc}
		\toprule
		\# of $\eta$ & TOF   & TOF ratio \\ \midrule
		2.5       & 67578 & 1.11      \\
		10        & \textbf{61098} & 1.00      \\
		20        & 93012 & 1.52      \\ \bottomrule
	\end{tabular}
\end{table}

\begin{table}[H]
	\caption{Sensivity to "Num adjust" (tested on \texttt{superblue19} and using cell inflation). TOF represents the sum of $OF(i,j,k)$}
	\label{tab:sens-adjust}
	\begin{tabular}{ccc}
		\toprule
		\# of   adjust & TOF           & TOF ratio \\ \midrule
		3              & 16862         & 2.14      \\
		4              & \textbf{5278} & 1.00      \\
		5              & 44000         & 5.58      \\ \bottomrule
	\end{tabular}
\end{table}

\begin{table}[H]
	\caption{Sensivity to $exponent$ (tested on \texttt{superblue19} and only using cell inflation). TOF represents the sum of $OF(i,j,k)$}
	\label{tab:sens-exp}
	\begin{tabular}{ccc}
		\toprule
		\# of   $exponent$ & TOF           & TOF ratio \\ \midrule
		0.5              & 14618         & 2.76     \\
		1.5              & \textbf{5278} & 1.00      \\
		2.5              & 5562         & 1.05      \\ \bottomrule
	\end{tabular}
\end{table}

\subsection{Sensivity to Message Function}
\label{sec:app-message}
In the message passing layers of RouteGNN, we use weighted summation of edge attributes (weight) for $\gV\rightarrow\gU$ and inner-product of edge attributes (product) for $\gC\rightarrow\gV$ because:
\begin{enumerate}
	\item \textbf{Efficiency}: In $\gV\rightarrow\gU$, the number of cells connected to a net $u$ is usually bigger than the number of nets connected to a cell $v$, and weight requires less computation than product.
	\item \textbf{Effectiveness}: In $\gC\rightarrow\gV$, we use the product, which is more informative than weight, to fuse information for cell representations.
\end{enumerate}
\begin{table}[H]
	\caption{Sensitivity to different message functions (tested on \texttt{superblue1/2/3/4/5/6/7/10/11/12/14/15}). The results are averaged across netlists on the cell level.}
	\label{tab:sens-message}
	\begin{tabular}{cccc}
		\toprule
		Variant     & pearson       & spearman      & kendall       \\ \midrule
		$\gV\rightarrow\gU$ sum     & 0.64          & 0.65          & 0.53          \\
		$\gC\rightarrow\gV$ sum     & 0.63          & 0.64          & 0.53          \\
		$\gV\rightarrow\gU$ product & 0.64          & 0.65          & 0.53          \\
		$\gC\rightarrow\gV$ weight  & 0.61          & 0.63          & 0.52          \\ \midrule
		default     & \textbf{0.64} & \textbf{0.66} & \textbf{0.53} \\ \bottomrule
	\end{tabular}
\end{table}
For $\gU\rightarrow\gV$ and $\gV\rightarrow\gC$, we simply use direct summation (sum) as the information has been collected. Table \ref{tab:sens-message} further evaluates other variants of message functions.

\subsection{Evaluating the Variability of Placement Results}
As the prediction quality of RouteGNN can be influenced by random factors such as initialization, we conduct evaluations on each circuit five times and report the mean and standard deviation of all metrics. The extended results are shown in Table \ref{tab:rep-ispd11-noinflat}, \ref{tab:rep-ispd2011}, \ref{tab:rep-dac12-noinflat} and \ref{tab:rep-dac12}.

\subsection{Ablation Study of RouteGNN Prediction Accuracy}
Table \ref{tab:dac12-train-random} compares the RoutePlacer performance using a well-trained RouteGNN—representing high prediction accuracy—with a version using a randomly initialized RouteGNN, representing low prediction accuracy. The results indicate that RoutePlacer with well-trained RouteGNN significantly outperforms its low-accuracy counterpart.

\end{document}